\documentclass[12pt,a4paper]{article}
\usepackage[left=2.54cm,top=2.54cm,right=2.54cm,bottom=2.54cm,bindingoffset=0.5cm]{geometry}
\usepackage{CJKutf8}
\usepackage{mathptmx} 
\usepackage{caption}
\usepackage{graphicx} 
\usepackage{subfigure}
\usepackage{adjustbox}
\usepackage{booktabs}
\usepackage{float}
\usepackage{listings} 
\usepackage{bm}
\usepackage{xcolor} 
\usepackage{hyperref}
\usepackage{comment}
\usepackage{natbib}
\usepackage{fancyhdr}
\title{
The realization of tones in spontaneous spoken Taiwan Mandarin: a corpus-based survey and theory-driven computational modeling
}
\author{Yuxin Lu$^{1}$ , Yu-Ying Chuang$^{2}$, R.Harald Baayen$^{3}$ \\ \ \\ \
$^{1}$ Quantitative Linguistics, Eberhard Karls Universität Tübingen,  \\
Tübingen, Germany \\
Email: yuxin.lu@uni-tuebingen.de \\
$^{2}$ Department of Taiwan Culture, Languages and Literature, \\
National Taiwan Normal University, Taipei, Taiwan \\
Email: yuying.chuang@ntnu.edu.tw \\
$^{3}$ Quantitative Linguistics, Eberhard Karls Universität Tübingen, \\
Tübingen, Germany \\
Email: harald.baayen@uni-tuebingen.de \\
} 
\date{\today}

\providecommand{\keywords}[1]
{
  \small	
  \textbf{\textit{Keywords:}} #1
}

\begin{document}
\begin{CJK*}{UTF8}{bsmi}

\maketitle

\newpage

\begin{abstract}

\noindent
A growing body of literature has demonstrated that semantics can co-determine fine phonetic detail. However, the complex interplay between phonetic realization and semantics remains understudied, particularly in pitch realization. The current study investigates the tonal realization of Mandarin disyllabic words with all 20 possible combinations of two tones, as found in a corpus of Taiwan Mandarin spontaneous speech. We made use of Generalized Additive Mixed Models (GAMs) to model f0 contours as a function of a series of predictors, including gender, tonal context, tone pattern, speech rate, word position, bigram probability, speaker and word. In the GAM analysis, word and sense emerged as crucial predictors of f0 contours, with effect sizes that exceed those of tone pattern. 
For each word token in our dataset, we then obtained a contextualized embedding by applying the GPT-2 large language model to the context of that token in the corpus. We show that the pitch contours of word tokens can be predicted to a considerable extent from these contextualized embeddings, which approximate token-specific meanings in contexts of use. The results of our corpus study show that meaning in context and phonetic realization are far more entangled than standard linguistic theory predicts.

\end{abstract}

 \\ \ \\ \ \\

\newpage

\section{Introduction}

\noindent
Mandarin Chinese is a tone language with four lexical tones: a high level tone (T1), a rising tone (T2), a low falling-rising tone known as a dipping tone (T3), and a falling tone (T4). Mandarin Chinese also has a so-called neutral or floating tone (T0), which is often described as unstressed, weaker in intensity, and shorter in duration \citep{chao1968grammar}.  The present study reports the results of an investigation of the realization of the Mandarin  tones in a corpus of Taiwan Mandarin spontanenous speech.  We first present our corpus based findings, and then present a theory-driven explanation of our findings using the Discriminative Lexicon Model \citep{Baayen:Chuang:Shafei:Blevins:2019, Heitmeier:Chuang:Baayen:2025}.

The corpus that we made use of was compiled by  \citet{fon_preliminary_2004}, originally with the aim of clarifying the influence of Southern Min on Mandarin Chinese as spoken in Taiwan. In what follows, we refer to this corpus as the Corpus of Spontaneous Taiwan Mandarin. The focus of our study is on the realization in this corpus of the tones in words consisting of two syllables.  The tonal realization of disyllabic words has been studied before in laboratory speech. \citet{xu1997contextual} examined the pitch contours of the 16 combinations of the 4 standard tones realized on the two-syllables /ma-ma/,  embedded in carrier sentences, and produced by male speakers of Beijing Mandarin.  Factors that are known to co-determine the realization of tones, such as speaking rate and the tones on adjacent words, were carefully controlled for.   This study showed that in laboratory speech, the tones of the single-syllable constituents are often somewhat different.  For instance, a rising tone followed by a falling tone (T2-T4) was observed to be realized as a fall, followed by a rise, and concluded with a fall. 

To our knowledge, there currently are no studies that address the tonal realizations of all tonal combination for disyllabic words in spontaneous conversation. It is well-known that spontaneous speech can differ markedly from formal speech.  Given that in spontaneous speech, words are often realized with various reduced forms \citep[see, e.g.,][for Dutch, English, and Mandarin Chinese, respectively]{Ernestus:2000,Johnson:2004,chung2006contraction}, it is an open question to what extent the canonical four tones of Mandarin are preserved in spontaneous speech.  This is one reason why we carried out a detailed investigation of the realization of tone in disyllabic words as found in the Corpus of Spontaneous Taiwan Mandarin.  Importantly, we considered not only the 16 combinations of tones studied by \citet{xu1997contextual}, but also the 4 combinations of a standard tone followed by a neutral tone (T1-T0, T2-T0, T3-T0, and T4-T0). 

The second reason we carried out this corpus survey is that previous corpus-based research provides strong evidence for the realization of words' tones, as represented by their f0 (pitch) contours, is only in part determined by the canonical tones of the constituent syllables, and that, surprisingly, words' meanings play a much more important role \citep{chuangwords,chuang2024word,lu2024sandhi,jin2024corpus} in shaping how the tones are actually realized. In section~\ref{sec:previousWork} we provide further details on these findings, and also point to several other studies indicating that meaning and phonetic form are far more entangled than is generally assumed.  Here, we note that if indeed fine semantic detail is reflected in fine phonetic detail, this challenges influential axioms of linguistic theory, such as the arbitrariness of the linguistic sign \citep{DeSaussure:66} and the dual articulation of language \citep{Martinet:65}.

A recent theory of the lexicon and lexical processing that rejects these axioms is the Discriminative Lexicon Model \citep{Baayen:Chuang:Shafei:Blevins:2019,Heitmeier:Chuang:Baayen:2025}.  This model represents both words' forms and their meanings as high-dimensional numeric vectors, and posits functions that map form vectors onto meaning vectors for comprehension, and meaning vectors onto form vectors (production). In the present study, we zoom in on only a small part of the production process, and ask whether it is possible to start out with a word's meaning vector (using context-specific embeddings from distributional semantics and Large Language Models) and to predict that word's pitch contour using a general mapping from semantic vectors to pitch contours.  In section~\ref{sec:modeling}, we show that this is indeed possible with an accuracy that is surprisingly far above chance level.  We will also show that the canonical tone pattern of a two-syllable word can be predicted from the centroid of the embeddings of the words sharing that canonical tone pattern. Our results raise many questions, for which, as will become clear in the general discussion, we only have tentative answers.  

The remainder of this paper is structured as follows. Section~\ref{sec:previousWork} introduces the many factors that co-determine how tones are realized, and also provides an overview of previous research on isomorphies between semantics and phonetic realization. Section~\ref{sec:corpus} introduces the corpus that we investigated, and provides details on data pre-processing and the statistical method that we used to analyze the corpus data.  Section~\ref{sec:results} reports our results: across all 20 tone patterns, words' meanings provide a surprisingly good window on their pitch contours.  Section~\ref{sec:modeling} presents our theory-driven computational modeling study, showing that token-specific pitch contours can be predicted from token-specific embeddings calculated based on their discourse context. Finally, Section \ref{sec:gendisc} presents our thoughts on the implications of our findings.

\section{Semantics and phonetic realization}\label{sec:previousWork}

\subsection{Spoken duration and articulation}

\noindent
Evidence is accumulating that subtle differences in meaning can be reflected in the fine phonetic details of how words are actually realized in corpora of natural speech, including aspects such as spoken word \citep{gahl2024time} duration, segment duration \citep{plag2017homophony}, and tongue position \citep{Saito_thesis}.  

Heterographic homophones are words with the same pronunciation but different spellings and meanings, such as \textit{time} and \textit{thyme}. For a long time, homophones were thought to sound identical \citep[see, e.g.,][]{Jescheniak:Levelt:94}. However,  \citet{gahl_time_2008}, using the Switchboard corpus \citep{godfrey1992switchboard} reported that heterographic homophones such as  \textit{time} and \textit{thyme} have different acoustic durations, with more frequent homophones (\textit{time}) being pronounced with shorter durations than their less frequent homophonic counterparts (\textit{thyme}). \citet{lohmann2018cut} similarly observed that the duration of words such as \textit{cut} depends on whether they are used as nouns or verbs. Both studies explain these effects in terms of how frequency of use affects lexical access in speech production. However, \citet{gahl2024time} reported that computational modeling with the Discriminative Lexicon Model provided strong evidence that the meanings of English homophones (represented by embeddings) are strong co-determinants of their spoken word duration, even when word frequency is controlled for. They argued that a powerful predictor of a homophone's spoken word duration is the degree of support it receives from the semantics, such that greater semantic support predicts longer spoken word duration. 

In addition to durational differences at the word level, durational differences have also been observed at the phonemic level in corpus studies, particularly for  the realization of word-final /s/ or /z/ (henceforth referred to as S) in English. In the Buckeye corpus \citep{pitt2005bcc}, word-final S has been found to vary in duration depending on its morphological function: non-morphemic S is pronounced longer than plural S, which, in turn, is pronounced longer than clitic S \citep{plag2017homophony, TOMASCHEK_PLAG_ERNESTUS_BAAYEN_2021,zimmermann2016morphological}. Furthermore, \citet{plag2020} found that genitive plural S showed significantly longer durations than plural S. \citet{schmitz2022production} utilized a pseudo-word paradigm to demonstrate that the morphological category of word-final S (non-morphemic $>$ plural $>$ clitics) influences its phonetic realization.

The relationship between semantics and phonetic realization has also been demonstrated beyond durational differences. \citet{drager2011sociophonetic} found that the pronunciation of the English word \textit{like} varies according to its discourse or grammatical meanings, not only in the duration of the consonants but also in the degree of diphthongization of the vowel. Furthermore, in line with \citet{gahl2024time}, \citet{Saito_thesis} and \citet{saito2024articulatory} reported for the KEC corpus of German spontaneous speech \citep{Arnold:Tomaschek:2016}, which also registers tongue movements using electromagnetic articulography, greater semantic support leads to a lower position of the tongue tip for the vowel /a/, indicating hyperarticulation. 

\subsection{Tone in Mandarin Chinese}

The preceding section reviewed recent evidence that semantics and phonetic realization are entangled to a greater extent than has often been assumed. In this section, we zoom in on how word meaning affects the realization of tone.  To do so, we first need to provide some further background on the factors that have already been reported to co-determine the realization of tone.

It is well known that the way in which tones are realized in connected speech differs from their canonical realization. How tones are realized has been described as depending on the properties of the segments in the syllable that carries a given tone \citep{ho1976mandarin, ohala1976explaining,xu2003effects}. Tonal variation in connected speech has also been reported to be shaped by the tones of adjacent words \citep[tonal coarticulation][]{xu1997contextual}, by speaking rate \citep{xu2002maximum}, by sentence intonation \citep{ho1976mandarin, wu:hal-03153402} and by a speaker's individual speaking style \citep{stanford2016sociotonetics}. 

At the socio-geographic level, cross-dialectal research has reported that different varieties of Mandarin exhibit varying tone inventories \citep{chang2010dialect, zhao2023production}.  For the present study, we note that in Standard Mandarin, the realization of a neutral tone following a given  lexical tone has been reported to be largely determined by this preceding tone. Furthermore,  the f0 contour of a neutral tone has been claimed to approach a low pitch target by the end of the carrying syllable \citep{xu2024cross}. However, in Taiwan Mandarin, the behavior of the the neutral tone has been reported to be different.  It can either be indistinguishable from one of the four canonical lexical tones, or it can be realized as a static mid-low pitch target \citep{huang2018phonological}.

From the above overview, it will be clear that the realization of tone is co-determined by a multitude of different factors. A newcomer in this arena is word meaning. \citet{chuang2024word} studied the pitch contours of di-syllabic words with an initial rising tone followed by a falling tone (henceforth the T2-T4 tone pattern). This study used a generalized additive model \citep[GAM][]{wood_generalized_2017} to decompose an observed pitch contour into separate pitch contours, capturing the effects over time of predictors such as speech rate, neighboring tones, and segmental properties. What \citet{chuang2024word} report is that the GAMs provide strong support for word-specific pitch contour components, while controlling for other variables such as segments, gender, speaker, speech rate, and the tones of adjacent words.  They also show that the statistical evidence is even stronger for sense-specific pitch contours, which suggests that these effects are semantic in nature.  The importance of words' meanings has been replicated for Mandarin disyllabic words with T2-T3 and T3-T3 tone pattern by \citet{lu2024sandhi}, and for monosyllabic Mandarin words by \citet{jin2024corpus}. For di-syllabic words with T2-T3 and T3-T3 tone pattern, the variable importance of words' meanings was on a par of that of tonal context (tone sandhi), the other most important predictor of words' pitch contours.  

\begin{figure}[htbp]
  \centering
  \subfigure[]{\includegraphics[width=2.75in]{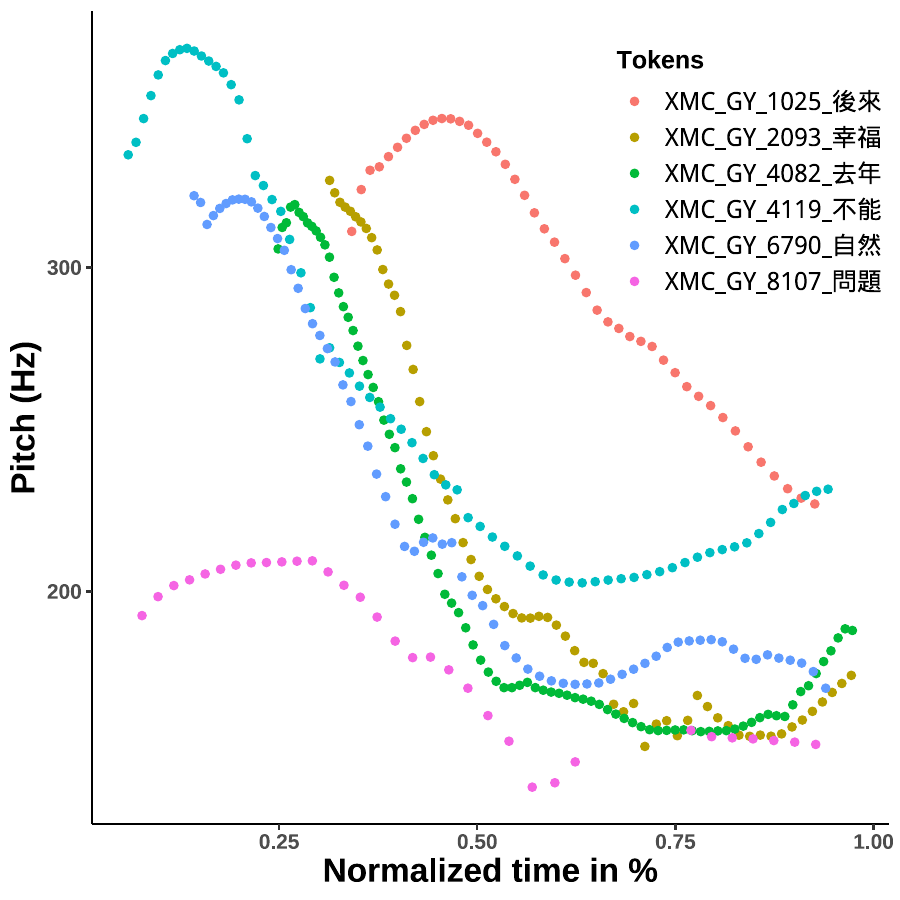}}
  \hspace{.25in}
  \subfigure[]{\includegraphics[width=2.75in]{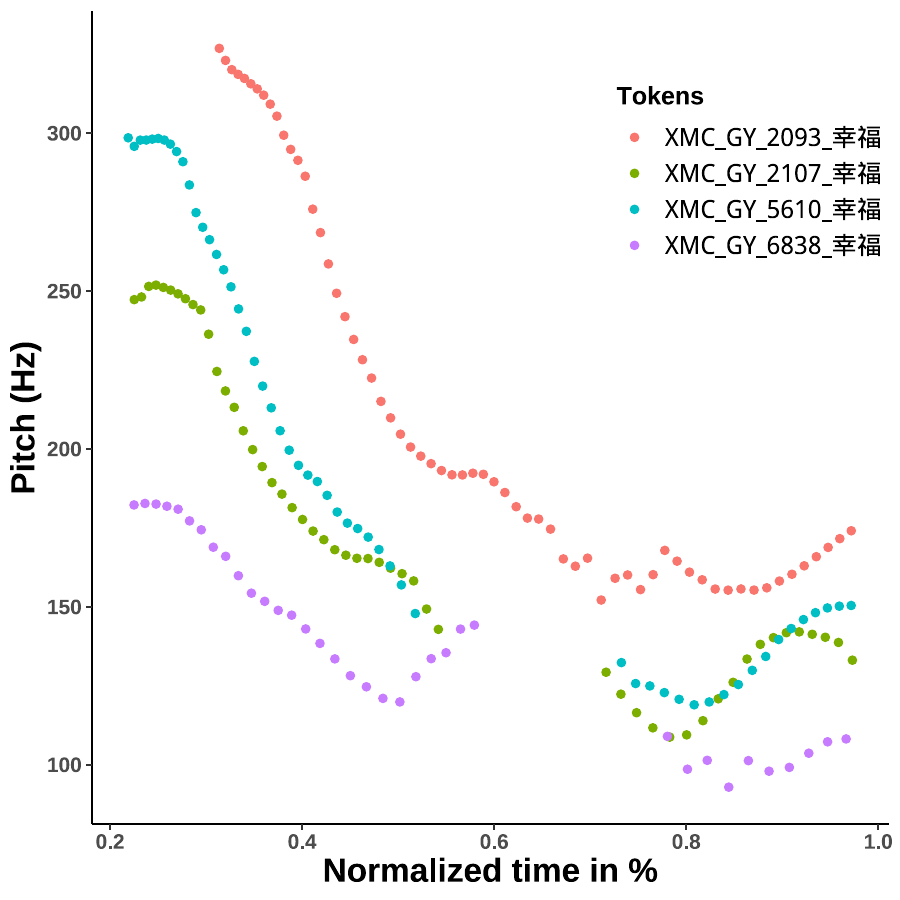}}
  \caption{A selection of tokens in spoken Taiwan Mandarin. 
  Left panel: six tokens representing six different word types, all sharing the tone pattern T4-T2 (a falling tone followed by a rising tone). The tokens are 後來\ (\textit{hou4lai2}, `later'), 幸福\ (\textit{xing4fu2}, `happiness'), 去年\ (\textit{qu4nian2},`last year'), 不能\ (\textit{bu4neng2},`cannot'), 自然\ (\textit{zi4ran2},`nature'), 問題\ (\textit{wen4ti2},`problem'). Right panel: four tokens representing the word type 幸福\ (\textit{xing4fu2}, `happiness'). All f0 contours shown here are produced by the same speaker.}
  \label{fig:toy}
\end{figure}

To illustrate the challenges that an analysis of tones in natural speech has to face, consider Figure~\ref{fig:toy}, which displays the f0 contours of a selection of tokens of word types with a falling tone followed by a rising tone (the T4-T2 tone pattern) in the Corpus of Spontaneous Taiwan Mandarin  \citep{fon_preliminary_2004}. In the left panel, the pitch contours of six tokens from different word types are presented. All words have the same canonical tone pattern: T4-T2.  Token  XMC\_GY\_4119\_不能\ (\textit{bu4neng2}, `cannot') (indicated by light blue) shows an initial sharp f0 rise, followed by a fall, and then a shallow rise.  Token XMC\_GY\_8107\_問題\ (\textit{wen4ti2}, `problem') (indicated by purple) has a much lower initial f0 than the other tokens. The initial f0 of token XMC\_GY\_1025\_後來\ (\textit{hou4lai2}, `later') (indicated by red) is unavailable due to the unvoiced initial /h/.  In the right panel of Figure~\ref{fig:toy}, we present four tokens of the same word type 幸福\ (\textit{xing4fu2}, `happiness'). One of its tokens is also presented in the left panel (indicated by yellow). The four tokens of 幸福\ also exhibit considerable variability in their f0 contours. 

In what follows, we take on the challenge of modeling the realization of tone, 
taking into account the many factors reported to co-determine pitch contours, such as gender, speaker, neighboring tones, speech rate, word position, and bigram probability.  Following up on earlier work \citep{chuang2024word, jin2024corpus, lu2024sandhi}, the `pitch signatures'  of individual words are of primary interest. Two hypotheses guide our research.
\begin{enumerate}
    \item The meanings of words co-determine the phonetic details of how the tones of these words are produced.
    \item The pitch contours of word tokens as found in spontaneous Mandarin conversations can be predicted from token-specific meaning vectors with above-chance accuracy using computational modeling. 
\end{enumerate}
In the next section, we describe the data that we collected from the Corpus of Spontaneous Taiwan Mandarin, and present our statistical analyses.  Section~\ref{sec:modeling} complements this exploratory part of our study with theory-driven computational modeling.

%

\section{Data collection and statistical analysis}\label{sec:corpus}

\subsection{The corpus}

\noindent
The data used in the present study come from the Taiwan Mandarin Spontaneous Conversation Corpus \citep{fon_preliminary_2004}. This corpus contains 30 hours of speech from 55 native speakers of Taiwan Mandarin, with 31 females and 24 males (aged between 20 and 60 years old). In unstructured interviews, participants were encouraged to speak freely, instead of being guided by a standardized set of questions. As a result, this corpus consists of naturally occurring speech with a diverse and varied set of words across speakers.

The corpus was transcribed in traditional Chinese characters at the word level. The speech data were segmented at both the syllable and word levels. Forced alignment was first performed, and the results were later manually reviewed by native Taiwan Mandarin speakers with a background in phonetics. In the current study, we followed the transcriptions and segmentation provided in the corpus.

\subsection{Data selection}

Disyllabic words with the 20 tone patterns were extracted for analysis (see column 1 in Table ~\ref{tab:counts}). The original dataset comprises 93,701 tokens, representing 7,526 unique word types. Table ~\ref{tab:counts} presents the counts of tokens and word types associated with each tone pattern. Among these, the T4-T4 pattern is the most frequent in disyllabic words, both token-wise and type-wise. The four tone patterns featuring a neutral tone in the second syllable (T1-T0, T2-T0, T3-T0, and T4-T0) are represented by the lowest numbers of types. There are also relatively few words with T3 \citep[see][for similar observations for journalistic speech]{wu2021tone}.

\begin{table}[htbp]
\centering
\caption{Number of tokens and words grouped by tone pattern in the conversational Taiwan Mandarin corpus.}
\label{tab:counts}
\begin{tabular}{rcrrl} \hline
 & \textbf{Tone pattern} & \textbf{Tokens} & \textbf{Word types} & \textbf{Examples} \\ 
  \hline
  1 & T1-T1 & 3501 & 459 & 應該\ (\textit{ying1gai1}, `should') \\
  2 & T1-T2 & 3725 & 458 & 當然\ (\textit{dang1ran2}, `of course') \\	
  3 & T1-T3 & 2313 & 333 & 根本\ (\textit{gen1ben3}, `at all') \\
  4 & T1-T4 & 7524 & 706 & 接觸\ (\textit{jie1chu4}, `to touch') \\
  5 & T1-T0 & 3034 &  83 & 他們\ (\textit{ta1men0}, `they') \\
  6 & T2-T1 & 2763 & 286 & 其他\ (\textit{qi2ta1}, `others') \\
  7 & T2-T2 & 3043 & 369 & 同學\ (\textit{tong2xue2}, `classmate') \\
  8 & T2-T3 & 4539 & 249 & 結果\ (\textit{jie2guo3}, `result') \\
  9 & T2-T4 & 9237 & 687 & 學校\ (\textit{xue2xiao4}, `school') \\
  10 & T2-T0 & 7010 &  64 & 什麼\ (\textit{shen2me0}, `what')\\
  11 & T3-T1 & 2655 & 252 & 老師\ (\textit{lao3shi1}, `teacher') \\
  12 & T3-T2 & 3465 & 289 & 感覺\ (\textit{gan3jue2}, `feeling') \\
  13 & T3-T3 & 3896 & 276 & 了解\ (\textit{liao3jie3}, `to know')\\
  14 & T3-T4 & 7256 & 595 & 可是\ (\textit{ke3shi4}, `but')\\
  15 & T3-T0 & 3295 &  50 & 我們\ (\textit{wo3men0}, `we')\\ 
  16 & T4-T1 & 3007 & 400 & 那些\ (\textit{na4xie1}, `those')\\ 
  17 & T4-T2 & 3978 & 451 & 後來\ (\textit{hou4lai2}, `later') \\
  18 & T4-T3 & 3302 & 419 & 父母\ (\textit{fu4mu3}, `parents')\\ 
  19 & T4-T4 & 13174 & 989 & 社會\ (\textit{she4hui4}, `society')\\ 
  20 & T4-T0 & 2984 & 111 & 爸爸\ (\textit{ba4ba0}, `daddy') \\ \hline
     & \textbf{Total} & 93701 & 7526 \\ \hline 
\end{tabular}
\end{table}

Subsequently, we extracted the sound files of these disyllabic words and measured their f0 values using the \textit{To Pitch (cc)} command in Praat \citep{boersma_praat_2020}. For female speakers, the pitch floor was set at 75 Hz and the pitch ceiling at 400 Hz. For male speakers, the pitch floor was set at 50 Hz and the pitch ceiling at 300 Hz. The time step was set to 0.001 seconds, and a Gaussian window was used for optimal F0 estimation. The \textit{To PointProcess} command was then applied to identify the time points of glottal pulses in the voiced sections, from which the corresponding F0 values were extracted. No f0 values were returned when there was no vocal fold vibration due to the presence of voiceless plosives or fricatives, or when creaky voice occurred.

Words with fewer than six tokens were excluded from our dataset, to ensure that each word type had a sufficient number of tokens for analysis. For high-frequency words with more than 200 tokens, we randomly sampled 200 tokens, to prevent model predictions from being biased towards high-frequency words.  Furthermore, words contributed by only female speakers or only by male speakers were excluded. This ensured that the tokens of a given word type were contributed by at least two speakers, preventing bias from one speaker’s specific way of speaking.

Lastly, tokens with f0 extraction errors were excluded from analysis. These errors typically resulted from pitch halving or doubling. We calculated, for each token, the standard deviation of the differences between consecutive measurements. A large standard deviation indicated high likelihood of discontinuous f0 measurements with abrupt fluctuations. Tokens with a standard deviation greater than the 9th decile of the distribution were considered outliers. 

\subsection{Predictors}

\noindent
The response variable of interest is f0. We log-transformed f0 to obtain a response variable that approximately follows a Gaussian distribution. As our interest is in production rather than comprehension, we did not make use of modifications of the logarithmic transformation such as the MEL or BARK scales, which are optimized for human perception.  The predictors for \texttt{log f0} are as follows. \\ \ \\

\hangafter 1
\hangindent 4em
\noindent
\textbf{\texttt{normalized\_t}} \space \space 
For each token, time was normalized between 0 and 1, enabling the modeling of tokens with varying durations on a common scale. Since f0 values were measured every 15 ms, tokens with longer durations have more measurements and, consequently, more data points within the [0,1] interval of normalized time. \\

\hangafter 1
\hangindent 4em
\noindent
\textbf{\texttt{gender}} \space \space A categorical variable with two levels—\texttt{female} and \texttt{male}. Due to physiological differences, female speakers generally produce speech at a higher pitch than male speakers. \texttt{gender} is included as a control variable. \\

\hangafter 1
\hangindent 4em
\noindent
\textbf{\texttt{speaker}} \space \space A factor with anonymized speaker identifiers as levels, required for fine-tuning differences in speakers' height of voice.  \\

\hangafter 1
\hangindent 4em
\noindent
\textbf{\texttt{tone\_pattern}} \space \space The tonal pattern of the token, as listed in the \textit{tone pattern} column in Table ~\ref{tab:counts}. \\

\noindent
\hangafter 1
\hangindent 4em
\textbf{\texttt{tonal\_context}} \space \space \texttt{preceding\_tone} is the tone of the syllable immediately preceding a token. \texttt{following\_tone} is the tone of the syllable immediately following a token. If a pause occurs immediately before or after the token, it is coded as \texttt{PAUSE}. Thus, both \texttt{preceding\_tone} and \texttt{following\_tone} include six possible values: 1, 2, 3, 4, 0, and PAUSE. To represent the different tonal contexts in which the token may appear, we define \texttt{tonal\_context} as the interaction of \texttt{preceding\_tone} and \texttt{following\_tone}, resulting in a factor with 36 levels. \\ 

\noindent
\hangafter 1
\hangindent 4em
\textbf{\texttt{speech\_rate}} \space \space Local speech rate, defined as the number of syllables per second for a given token, is calculated over a window extending four characters to the left and four characters to the right of the token. This measurement of speech rate is included as a covariate to control for potential effects of durational differences. To avoid concurvity, duration is therefore not included alongside \texttt{speech\_rate} as a predictor, as the two variables are moderately correlated $r = -0.55$. \\

\noindent
\hangafter 1
\hangindent 4em
\textbf{\texttt{norm\_utt\_pos}} \space \space 
Normalized position in the utterance represents the relative position of a word within its utterance. It is calculated by dividing the word's position by the total number of syllables in the utterance, resulting in a value normalized on a scale from 0 to 1. Higher values indicate that the token occurs closer to the end of the utterance. For single-word utterances, the position is coded as 1. Previous research has shown that utterance-final words tend to exhibit a rising pitch \citep{shih2000declination}. \\

\noindent
\hangafter 1
\hangindent 4em
\textbf{\texttt{bg\_prob\_prev}} \space \space 
Bigram probability quantifies how predictable a word is in its context. This measure of contextual predictability is based on the relative frequency of the word's co-occurrence with surrounding words. A higher bigram probability indicates that the target word is more predictable within its given context. In general, higher predictability is associated with shorter word durations and greater spectral reduction \citep{arnon2014time,tang2018contextual}. There is also some evidence showing that contextual predictability influences f0 production, as observed in English \citep{turnbull2017role}, Taiwan Mandarin \citep{hsieh2013prosodic}, and Taiwan Southern Min \citep{wang2024contrast}. In the present study, following \citet{gahl_time_2008}, \texttt{bg\_prob\_prev} is calculated as the probability of the occurrence of the target word given the preceding word. \\

\noindent
\hangafter 1
\hangindent 4em
\textbf{\texttt{bg\_prob\_fol}} \space \space This measure represents the bigram probability of the following word, calculated as the probability of the occurrence of the target word given the following word. \\

\noindent
\hangafter 1
\hangindent 4em
\textbf{\texttt{word}} \space \space A factor with orthographic words, as available in the corpus, as levels. For instance, the token XMC\_GY\_8107\_問題 \ is coded as 問題 \ using traditional Chinese characters. The dataset contains 313 unique words, so there are 313 corresponding levels for \texttt{word}.  \\

\noindent
\hangafter 1
\hangindent 4em
\textbf{\texttt{sense\_type}} \space \space 
A word can have multiple senses, which are identified based on the contexts in which the word occurs. We used a word sense identification system, described in \citet{hsieh_resolving_2024}, that utilizes BERT in combination with the Chinese WordNet \citep{huang_chinese_2010}. \\ \ \\


%

\noindent
Of the above list of predictors, the factor \texttt{tonal\_context} poses a special challenge for the analysis. \texttt{tonal\_context} provides information about the preceding and following tones.  Due to the pervasiveness of tonal co-articulation, it is highly probable that the effect of \texttt{tonal\_context} varies with the tone pattern of the target word. For example, a preceding high tone will have an effect on a word-initial dipping tone that differs from its effect on a word-initial rising tone. Accounting for such co-articulation is essential for modeling f0 in connected speech. In principle, one could introduce a variable that represents the interaction between \texttt{tonal\_context} and \texttt{tone\_pattern} \citep[cf.][]{jin2024corpus}. However, for our dataset, this would result in a variable with 720 levels that is strongly confounded with \texttt{word} and \texttt{sense\_type}.

Therefore, we opted to fit separate regression models for f0 across four different most frequent tonal contexts in our dataset, excluding any contexts that involved a ``pause'' in the preceding or following tone, i.e., the contexts \texttt{4.4}, \texttt{3.4}, \texttt{4.1}, and \texttt{4.0} (cf. Table~\ref{tab:dataset}). We chose not to include tonal contexts involving a ``pause'', for two reasons. First, when a tone is preceded or followed by a pause, several context-related variables, such as \texttt{norm\_utt\_pos}, \texttt{bg\_prob\_prev}, and \texttt{bg\_prob\_fol}, are missing, leading to data loss. Second, pauses in speech often signal utterance boundaries, hesitations, or breaths, making the ``pause'' category inherently heterogeneous.

As shown in Table~\ref{tab:dataset}, the final dataset contains 4,283 tokens representing 313 unique word types. The minimum number of tokens per word type is 5, and the maximum is 56. On average, each word type was produced by 9.37 different speakers (range: 2 to 30). Additionally, each speaker contributed an average of 53.31 different word types (range: 4 to 119). For each tonal context, all 20 tone patterns are represented.

\begin{table}[htbp]
\caption{Overview of the four sub-datasets grouped by the four tonal contexts.}
\centering
\adjustbox{max width=\textwidth}{
\begin{tabular}{c c c c c c} \hline
\textbf{Tonal context} & \textbf{Number of tokens} & \textbf{Number of word types} & \textbf{Number of tone patterns}\\ \hline
\texttt{4.4}  & 1,794 & 288 & 20 \\
\texttt{3.4}  & 888 & 240 & 20 \\
\texttt{4.1}   & 874 & 250 & 20 \\
\texttt{4.0}  & 727 & 210 & 20 \\ \hline
\textbf{Total} & 4,283 & 313 & 20 \\ \hline
\end{tabular}
}
\label{tab:dataset}
\end{table}

\subsection{Statistical analysis} \label{sec:results}

\subsubsection{Models with word as predictor} 

\noindent
The Generalized Additive Model \citep[GAM,][]{wood_generalized_2017} was used for the statistical analyses, with the \texttt{bam()} function from \texttt{mgcv} package \citep{wood_generalized_2017} implemented in R \citep{team_r_2020}. Four GAMs were fitted to each of the four datasets, using the same model specification:  

\begin{tabbing}
mmmmm\=mm\=\kill
\texttt{logf0} \> $\sim$ \> \texttt{gender} + \\
      \> \> \texttt{s(normalized\_t, by=gender,k=4) +} \\
       \> \> \texttt{s(speaker, bs=`re') +} \\
       \> \> \texttt{s(normalized\_t, tone\_pattern, bs=`fs', m=1)+} \\
       \> \> \textbf{s(normalized\_t, word, bs=`fs', m=1)} + \\
       \> \> \texttt{s(speech\_rate, by=gender, k=4) +} \\
       \> \> \texttt{ti(normalized\_t, speech\_rate, k=c(4,4)) +} \\
       \> \> \texttt{s(norm\_utt\_pos, k=4) +} \\
       \> \> \texttt{ti(normalized\_t, norm\_utt\_pos, k=c(4,4)) +} \\
       \> \> \texttt{s(bg\_prob\_prev, k=4)+} \\
       \> \> \texttt{ti(normalized\_t, bg\_prob\_prev, k=c(4,4))+} \\
       \> \> \texttt{s(bg\_prob\_fol, k=4)+} \\
       \> \> \texttt{ti(normalized\_t, bg\_prob\_fol, k=c(4,4))}
\end{tabbing}

To account for differences in average pitch height between genders, we included \texttt{gender} as a fixed effect. We added a by-gender thin plate regression smooth of \texttt{normalized\_t}, which allows us to capture differing relationships between normalized time and f0 across genders. Other continuous variables, including \texttt{speech\_rate}, \texttt{norm\_utt\_pos}, \texttt{bg\_prob\_prev}, and \texttt{bg\_prob\_fol} were likewise modeled with thin plate regression splines. Interactions of covariates with normalized time were modeled with tensor product smooths, using the \texttt{ti()} function. 

Furthermore, random intercepts were requested for \texttt{speaker} to account for individual variability in pitch height by speaker. Other discrete variables, including \texttt{tone\_pattern} and \texttt{word}, were modeled using factor smooths (nonlinear random effects). 

We implemented an AR(1) process (first-order auto-regressive model) in the residuals to take into account the auto-correlations in the time series of pitch measurements. The inclusion of the AR(1) process with an auto-correlation coefficient of rho = 0.95 effectively removed nearly all autocorrelation from the residuals.  Summaries of the four models are provided in the Appendix.

\begin{figure}[htbp]
  \centering
  \includegraphics[width=\textwidth]{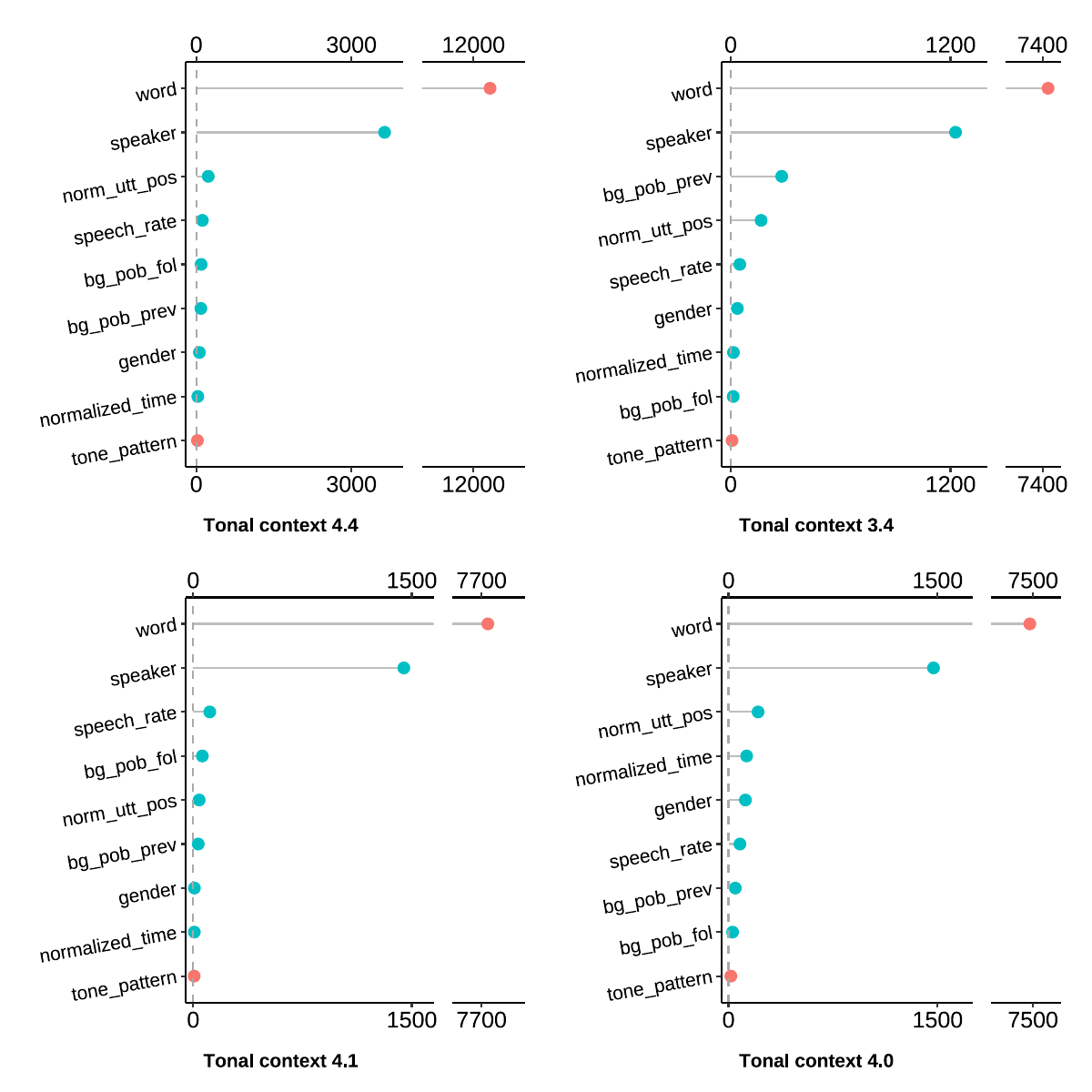}
  \caption{The increase in AIC scores when a predictor is withheld from the best-fit model. The AIC increase when \texttt{word} or \texttt{tone\_pattern} is withheld is shown in red, and the increase for other predictors is shown in blue. Panels 1 to 4 represent four GAMs with tonal contexts \texttt{4.4}, \texttt{3.4}, \texttt{4.1}, and \texttt{4.0}, respectively.}
  \label{fig:aic}
\end{figure}

Akaike's Information Criterion (AIC) was used to assess variable importance. Figure~\ref{fig:aic} shows the increase in AIC (indicating a lower-quality fit to the data) resulting from  withholding individual predictors from the model specification. A greater increase in AIC when a predictor is excluded suggests a higher importance of that predictor in the model. As shown in Figure~\ref{fig:aic}, across all tonal contexts, withholding the predictor \texttt{word} leads to a substantial increase in AIC scores, ranging from 7430.30 to 12320.26. The increase in AIC when \texttt{word} is omitted from the model specification substantially exceeds the corresponding change observed for any other predictor.

Surprisingly,  withholding \texttt{tone\_pattern} has a small impact on the model fit with increases in AIC of around 22.66 units (22.66 units for \texttt{4.4}, 7.33 units for \texttt{3.4}, 6.34 units for \texttt{4.1}, and  16.9 units for \texttt{4.0}). One possible explanation is that \texttt{word} is nested within \texttt{tone\_pattern}. When \texttt{word} is removed from the best-fit GAM, withholding \texttt{tone\_pattern} results in a larger AIC increase by 7354.08 units for \texttt{4.4},  4069.89 units for \texttt{3.4},  3026.09 units for \texttt{4.1} and  3451.28 units for \texttt{4.0}. This suggests that \texttt{tone\_pattern} still contributes to the model fit, though not as strongly as word. When word is included in the model, the effect of \texttt{tone\_pattern} is overshadowed by the stronger effect of word.

\begin{figure}[htbp]
  \centering
  \includegraphics[width=\textwidth]{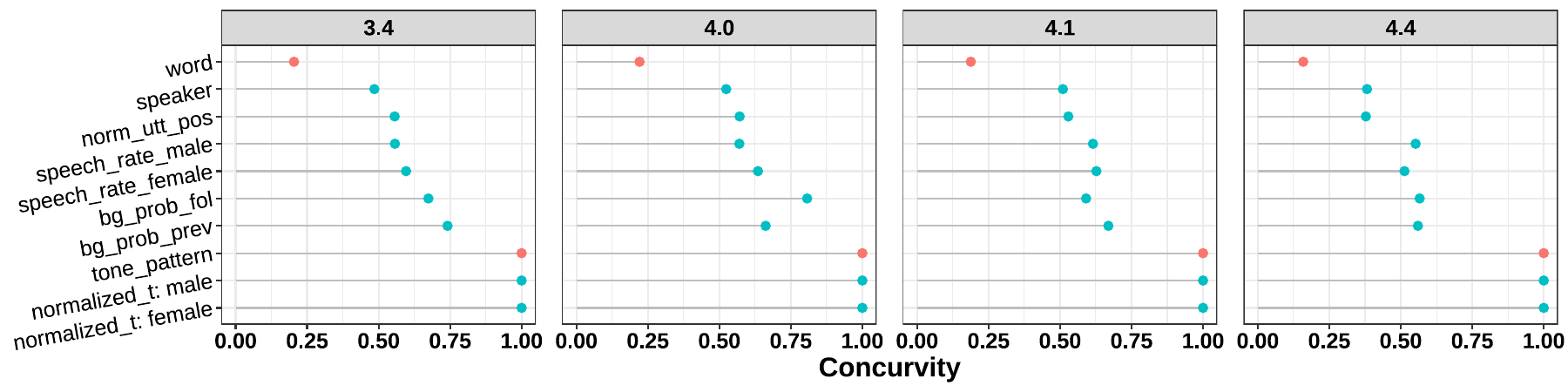}
  \caption{Concurvity scores for selected terms in the four GAMs. 
  The concurvity scores for \texttt{word} and \texttt{tone\_pattern} are shown in red, and those for other predictors are shown in blue. 
  From left to right, it presents tonal context \texttt{3.4}, \texttt{4.0}, \texttt{4.1}, and \texttt{4.4} respectively. Concurvity scores were calculated based on the best-fit GAMs with all predictors included.}
  \label{fig:concurvity}
\end{figure}

Concurvity, analogous to collinearity in linear regression, measures how much a predictor's effect can be explained by other predictors in the model. Concurvity scores range from 0 to 1, with lower values indicating that the contribution of a  predictor is less confounded with the contributions of other predictors. As shown in  Figure~\ref{fig:concurvity}, concurvity scores follow a similar pattern for all four GAMs, with the lowest concurvity scores for \texttt{word}. \texttt{speaker} also has relatively low concurvity.

By contrast, the predictor \texttt{tone\_pattern} exhibits extremely high concurvity, ranging from 0.998 to 1. This is due to tone pattern being fully predictable given the word.  When \texttt{word} is excluded from the model, the concurvity of \texttt{tone\_pattern} drops substantially to 0.09. This indicates that \texttt{word} captures information about the word's tone pattern, so when \texttt{word} is included in the model, the tone pattern is also included implicitly. However, if only \texttt{tone\_pattern} is specified, word-specific information is not available. This results in a substantially worse fit, which aligns with the AIC change discussed in preceding subsection.

\begin{figure} [htbp]
    \centering
    \includegraphics[width=\linewidth]{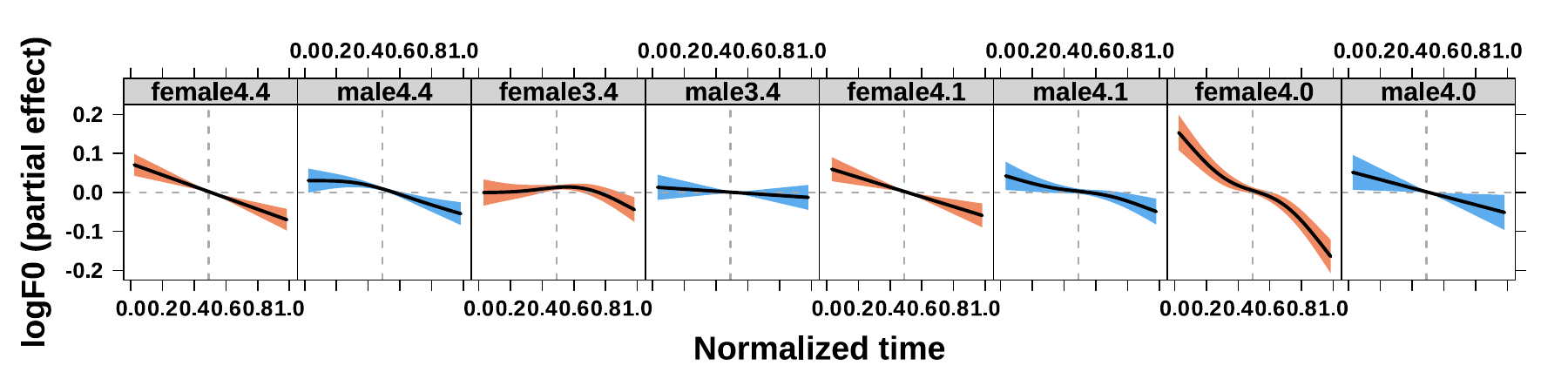}
    \caption{The partial effect of general smooth for the \texttt{normalized\_time} for female and male speakers, in different tone contexts. The orange curves indicate the general contours for female speakers, and the blue curves indicate the general contours for male speakers. Vertical grey dashed lines indicate the average syllable boundary, and the horizontal grey dashed line represents the y=0 reference line.}
    \label{fig:time}
\end{figure}

{\color{black}Finally, we note that the by-gender smooths for \texttt{time} (\texttt{normalized\_t:\-female} and \texttt{normalized}\linebreak \texttt{\_t:male})} show very high concurvity — unsurprisingly, as the tonal contours for both genders are highly similar (see Figure~\ref{fig:time}). Without accounting for other effects, these general contours primarily reflect the pure influence of time on pitch, illustrating how pitch contours change over time. The overall curves exhibit falling contours, which suggests a general declination trend in pitch contours for disyllabic words.

%
%
%

\begin{figure}[htbp]
  \centering
  \includegraphics[width=\textwidth]{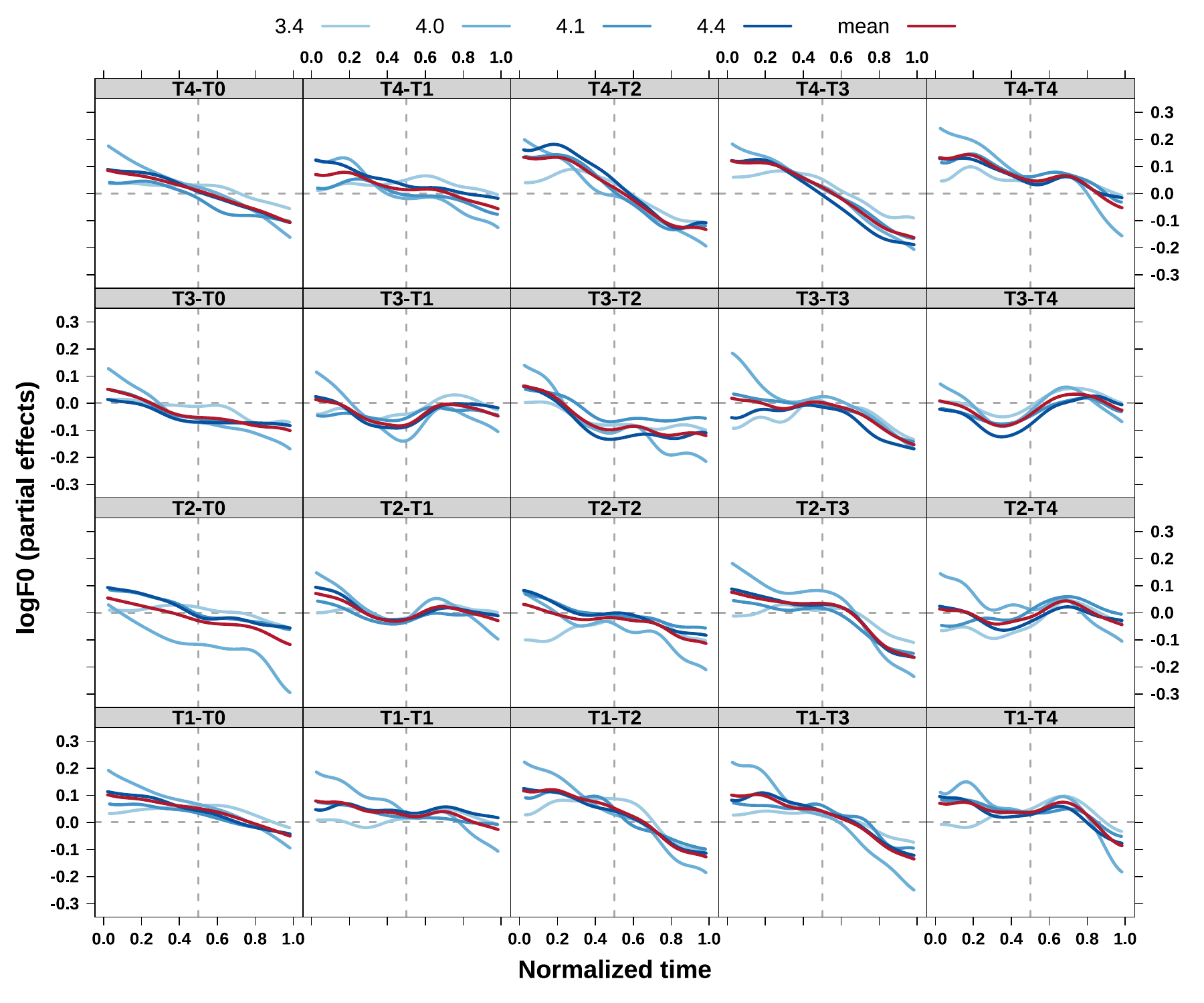}
  \caption{The effect of tone pattern. 
  The blue curves represent the partial effects of the factor smooth for \texttt{tone\_pattern}, combined with the general smooth of \texttt{normalized\_t} for female speakers, based on the best-fit models that include the word effect. There is one GAM for each tonal context, resulting in four blue curves representing, in a given panel, the four tonal contexts. The red curves present the mean f0 contours of a tone pattern, calculated by averaging the four f0 contours across the  tonal contexts. Thus, the blue curves in each panel illustrate how the tonal context modulates the general curve shown in red. Vertical grey dashed lines indicate the average syllable boundary, and the horizontal grey dashed line represents the y=0 reference line.}
  \label{fig:pattern}
\end{figure}

Figure~\ref{fig:pattern} illustrates the partial effects of the 20 tonal patterns across the four tonal contexts under investigation, using color coding to distinguish between the tonal contexts. Within each panel, the various blue curves represent specific tone patterns associated with different tonal contexts. For example, the lightest blue curve in the upper-left panel represents the T4-T0 tone pattern in the \texttt{3.4} tonal context. In other words, it represents a tonal sequence T3-T4-T0-T4, as in the phrase 有這麼重\  (\textit{you3zhe4me0zhong4}, `...is this heavy'). The red curves were obtained from averaging the four blue curves representing tone patterns under different tonal contexts. The deviations of the blue curves from the corresponding red curves highlight how the actual realizations of a tonal pattern in context differ from the expected effect of tone pattern, irrespective of context.   


For most of the tone patterns, the effects of the neighboring tones on the pitch contour are relatively modest, with as glaring exceptions the T2-T0 tone patterns in the \texttt{4.0} tonal context. This tonal sequence T4-T2-T0-T0 (e.g., 對孩子的, \textit{dui4hai2zi0de0}, ``for children's \ldots'') shows an unexpectedly low f0. This is probably due to the fact that this tonal sequence is underrepresented in the dataset, with only 9 tokens representing 4 unique word types (cf. Table~\ref{tab:counts} in Appendix 1. ).

However, the effects of tonal context are less pronounced in tone patterns featuring the neutral tone. For tone patterns T1-T0, T2-T0, T3-T0, and T4-T0, the general trend appears to approach a similar mid-low pitch target at the end of the syllable, regardless of the following tone.

For the \texttt{3.4} tonal context, 14 of the tonal patterns begin with the lowest f0. This may be a straightforward consequence of Tone 3 being often realized as a low tone in Taiwan Mandarin \citep{Fon1999does}. For the \texttt{4.0} tonal context, by the end of the word, the f0 tends to be the lowest across all panels. This is probably due to the general curve of female speakers in \texttt{4.0} tonal context. The female curve of \texttt{4.0} tonal context has a particularly salient falling trend (cf. Figure~\ref{fig:time}). Apparently, the following neutral tone is magnifying the final downward inclination observed in the vast majority of tone patterns.

%
%
%

\begin{figure}[htbp]
  \centering
  \includegraphics[width=\textwidth]{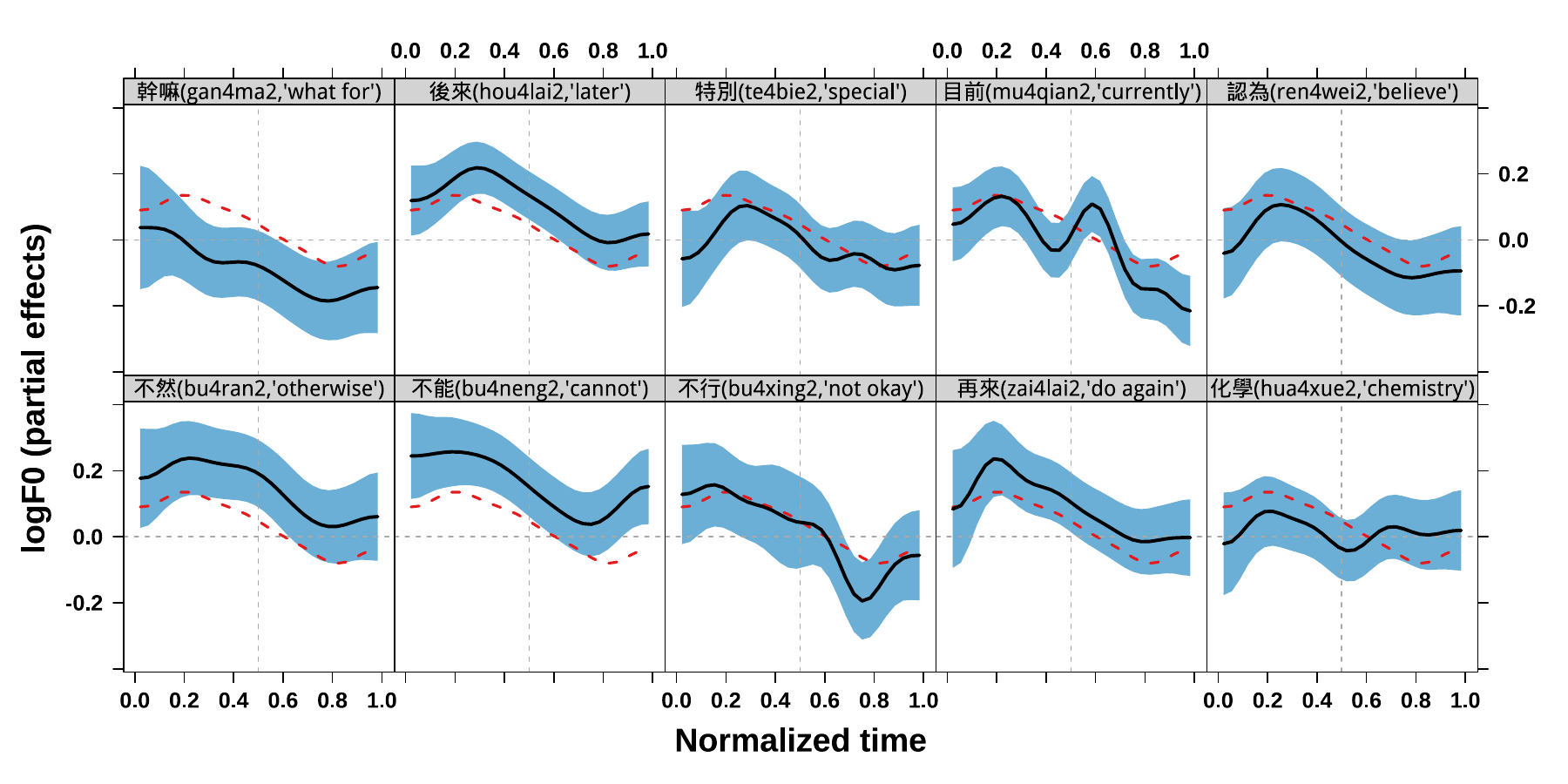}
  \caption{A selection of predicted f0 contours for words with the T4-T2 tone pattern. These contours are estimated by combining the partial effects of the factor smooth for \texttt{word} and the corresponding factor smooth for \texttt{tone\_pattern} (T4-T2). 
  The dashed red curve represents the partial effect of the T4-T2 tone pattern alone, which is identical across all panels.
  Vertical grey dashed lines indicate the average syllable boundary, while the horizontal grey dashed line represents the y=0 reference line.}
  \label{fig:word_pattern}
\end{figure}

Figure~\ref{fig:word_pattern} displays a selection of predicted pitch contours estimated by the factor smooth for \texttt{word}, combined with the partial effects of the factor smooth for \texttt{tone\_pattern}. All words presented here follow the T4-T2 tone pattern in the \texttt{4.4} tonal context (i.e.,  a tonal sequence T4-T4-T2-T4). For instance, this sequence occurs in the phrase 就變成興趣\ (\textit{jiu4bian4cheng2xing4qu4}, `then become an interest'). These partial effects exclude the general intercept and do not account for pitch differences between female and male speakers. 

The red dashed curves represent the partial effect of the factor smooth for \texttt{tone\_pattern} only, without incorporating the word-specific pitch contours, and are shown to provide a reference for assessing the word-specific effects. 

It can be observed that the pitch contours of 後來\ (\textit{hou4lai2}, `later'), 不然\ (\textit{bu4ran2}, `otherwise'), and 不能\ (\textit{bu4neng2}, `cannot') closely align with the predicted tone pattern but are overall shifted upward. Similarly, the pitch contour of 認為\ (\textit{ren4wei2}, `to believe') also follows a similar shape but is shifted downward. However, other words, such as 幹嘛\ (\textit{gan4ma2}, `What for?'), 目前\ (\textit{mu4qian2}, `at present'), and 化學\ (\textit{hua4xue2}, `chemistry'), largely deviate from the general tone pattern. Two words beginning with 不\ (\textit{bu4}, expressing negation), 不然\ (\textit{bu4ran2}, `otherwise') and 不能\ (\textit{bu4neng2}, `cannot'), have very similar contours that run parallel to the contour of the T4-T2 pattern. However, 不行\ (\textit{bu4xing2}, `not okay'), displays a steeper fall.

The deviation of the blue curves from the red dashed curves reflects the differences between the predicted pitch contours and the general tone pattern. The word-specific tonal realizations observed here are similar to those reported for Mandarin disyllabic words with T2-T4 tone patterns \citep{chuang2024word}, as well as words with the T2-T3 and T3-T3 tone patterns \citep{lu2024sandhi}.


\subsection{Sense-specific tonal realization} 

In the preceding section, we documented that the tonal realization of Mandarin di-syllabic words varies systematically by word.  It is possible that words'  segmental make-up is the crucial factor. Alternatively, it is theoretically possible that it is words' meanings that shape their pitch contours, just as in English, the duration of homophones is to a considerable extent co-determined by their meanings \citep{gahl2024time}.  If this hypothesis is on the right track, then word sense should be a more precise predictor than word identity. In the following  analysis, we explore whether we can replicate previous studies in which sense emerged as an even better predictor of disyllabic words' pitch contours than the word itself \citep{chuang2024word,lu2024sandhi}. If we can show that a word with different meanings exhibits varying pitch realizations, this will provide further evidence that words' meanings co-determine tonal realization. 

In order to explore the value of this hypothesis, we make use of the fact that our data are taken from a corpus, and not a word list.  As a consequence, we can estimate a word token's most likely sense in the exact context in which it was used. To determine these most likely senses in context, we made use of the sense identification system proposed by \citet{hsieh_resolving_2024}, which uses BERT in combination with the Chinese WordNet \citep{huang_chinese_2010}. For example, this system identifies the word 先生\ (\textit{xian1sheng1}, `husband, sir') as `a woman's spouse in a marital relationship' in sentences such as 我先生認為\ (\textit{wo3xian1sheng1ren4wei2}, `My husband thinks \ldots') or 我先生去睡覺\ (\textit{wo3xian1sheng1qu4shui4jiao4},`My husband went to sleep \ldots'). It assigns 先生\ (\textit{xian1sheng1}, `husband, sir') the sense `a man addressed in a social context' to when it appears in the phrase 那位先生\ (\textit{na4wei4xian1sheng1}, `That gentleman over there \ldots').

Since not all words in the dataset could be assigned a sense, we excluded words for which no sense type was identified. Second, we removed sense types represented by fewer than six tokens to ensure that each sense type had sufficient data for meaningful analysis. To prevent the model's predictions from being biased toward high-frequency sense types, we limited the maximum number of tokens per sense type. Specifically, for any sense type represented by more than 50 tokens, we randomly sampled 50 tokens from all tokens. We then grouped the dataset by tonal context, as in the previous analysis, resulting in four sub-datasets (see Table ~\ref{tab:sense_data}). The final dataset consists of 3,525 tokens representing 290 unique sense types. After the trimming process, 252 unique word types remain from the initial 313. All 20 tone patterns are present for each tonal context. The distribution of sense types, and word types follow the similar pattern as in the dataset shown in Table \ref{tab:counts}.

\begin{table}[htbp]
\caption{Overview of trimmed datasets grouped by the four tonal contexts, for the sense analysis.}
\centering
\adjustbox{max width=\textwidth}{
\begin{tabular}{l r r r r} \hline
\textbf{Tonal context} & \textbf{Tokens} & \textbf{Sense types} & \textbf{Word types} & \textbf{Tone patterns} \\ \hline
\texttt{4.4}   & 1512 & 266 & 233 & 20 \\
\texttt{3.4}   & 740  & 220 & 195 & 20 \\
\texttt{4.1}   & 716  & 228 & 200 & 20 \\
\texttt{4.0}   & 557  & 171 & 157 & 20 \\ \hline
\textbf{Total}  & 3525 & 290 & 252 & 20 \\ \hline
\end{tabular}}
\label{tab:sense_data}
\end{table}

\noindent
For the sense analysis, we replaced the factor smooth for \texttt{word} with a factor smooth for \texttt{sense\_type}, while keeping all other predictors from the previous analysis.

\begin{tabbing}
mmmmm\=mm\=\kill
\texttt{logf0} \> $\sim$ \> \texttt{gender} + \\
      \> \> \texttt{s(normalized\_t, by=gender,k=4) +} \\
       \> \> \texttt{s(speaker, bs='re') +} \\
       \> \> \texttt{s(normalized\_t, tone\_pattern, bs=`fs', m=1)+} \\
       \> \> \textbf{s(normalized\_t, sense\_type, bs=`fs', m=1)} + \\
       \> \> \texttt{s(speech\_rate, by=gender, k=4) +} \\
       \> \> \texttt{ti(normalized\_t, speech\_rate, k=c(4,4)) +} \\
       \> \> \texttt{s(norm\_utt\_pos, k=4) +} \\
       \> \> \texttt{ti(normalized\_t, norm\_utt\_pos, k=c(4,4)) +} \\
       \> \> \texttt{s(bg\_prob\_prev, k=4)+} \\
       \> \> \texttt{ti(normalized\_t, bg\_prob\_prev, k=c(4,4))+} \\
       \> \> \texttt{s(bg\_prob\_fol, k=4)+} \\
       \> \> \texttt{ti(normalized\_t, bg\_prob\_fol, k=c(4,4))}
\end{tabbing}

\noindent
An AR(1) process in the errors was also included to account for the autocorrelation in the pitch time series. The model summary is available in the Appendix.

To assess the relative importance of \texttt{sense\_type}, \texttt{word}, and \texttt{tone\_pattern}, we compared three additional models with different predictor structures: (1) a model with \texttt{sense\_type} + \texttt{tone\_pattern}, (2) a model with \texttt{word} + \texttt{tone\_pattern}, and (3) a model \texttt{sense\_type} by itself. Table ~\ref{tab:aic_sense} presents the AIC differences resulting from changing or withholding the given variable, relative to the (\texttt{sense\_type} + \texttt{tone\_pattern}) model. 

First consider the GAMs where \texttt{sense\_type} was replaced by \texttt{word}. In the case of the \texttt{4.4} tonal context, replacing \texttt{sense\_type} with \texttt{word} (comparing row 1 and row 2) led to a substantial AIC increase of 457.28 units. This suggests that \texttt{sense\_type} is a stronger predictor than \texttt{word} for modeling f0 contours. 

Second, comparing row 1 and row 3, removing \texttt{tone\_pattern} while retaining \texttt{sense\_type} led to a smaller AIC increase of 28.08 units. This indicates that \texttt{tone\_pattern} contributes to the model fit, albeit with a relatively minor effect when \texttt{sense\_type} is included. A similar AIC pattern across the \texttt{3.4}, \texttt{4.1}, and \texttt{4.0} tonal contexts further reinforces the stronger influence of \texttt{sense\_type} over \texttt{word} in modeling f0 contours.

\begin{table}[htbp]
    \centering
    \caption{AIC scores for models with different structures of \texttt{word}, \texttt{sense\_type}, and \texttt{tone\_pattern}, fitted separately to datasets for the four tonal contexts.}
    \adjustbox{max width=\textwidth}{
    \begin{tabular}{lcccc} \hline
    \textbf{Tonal Context} & \textbf{Model} & \textbf{AIC} & \textbf{AIC Difference} \\ \hline
    4.4 & all other predictors + \texttt{sense\_type} + \texttt{tone\_pattern} & -226847.29 & – \\
    4.4 & all other predictors + \texttt{word} + \texttt{tone\_pattern} & -226390.01 & 457.28 \\
    4.4 & all other predictors + \texttt{sense\_type} & -226824.40 & 22.89 \\ \hline
    3.4 & all other predictors + \texttt{sense\_type} + \texttt{tone\_pattern} & -117518.90 & – \\
    3.4 & all other predictors + \texttt{word} + \texttt{tone\_pattern} & -116923.04 & 595.85 \\
    3.4 & all other predictors + \texttt{sense\_type} & -117515.43 & 3.47 \\ \hline
    4.1 & all other predictors + \texttt{sense\_type} + \texttt{tone\_pattern} & -113771.84 & – \\
    4.1 & all other predictors + \texttt{word} + \texttt{tone\_pattern} & -113177.81 & 594.03 \\
    4.1 & all other predictors + \texttt{sense\_type} & -113765.18 & 6.66 \\ \hline
    4.0 & all other predictors + \texttt{sense\_type} + \texttt{tone\_pattern} & -93868.50 & – \\
    4.0 & all other predictors + \texttt{word} + \texttt{tone\_pattern} & -93650.04 & 218.46 \\
    4.0 & all other predictors + \texttt{sense\_type} & -93861.31 & 7.20 \\ \hline
    \end{tabular}}
    \label{tab:aic_sense}
\end{table}

\begin{figure}[htbp]
  \centering
  \includegraphics[width=\textwidth]{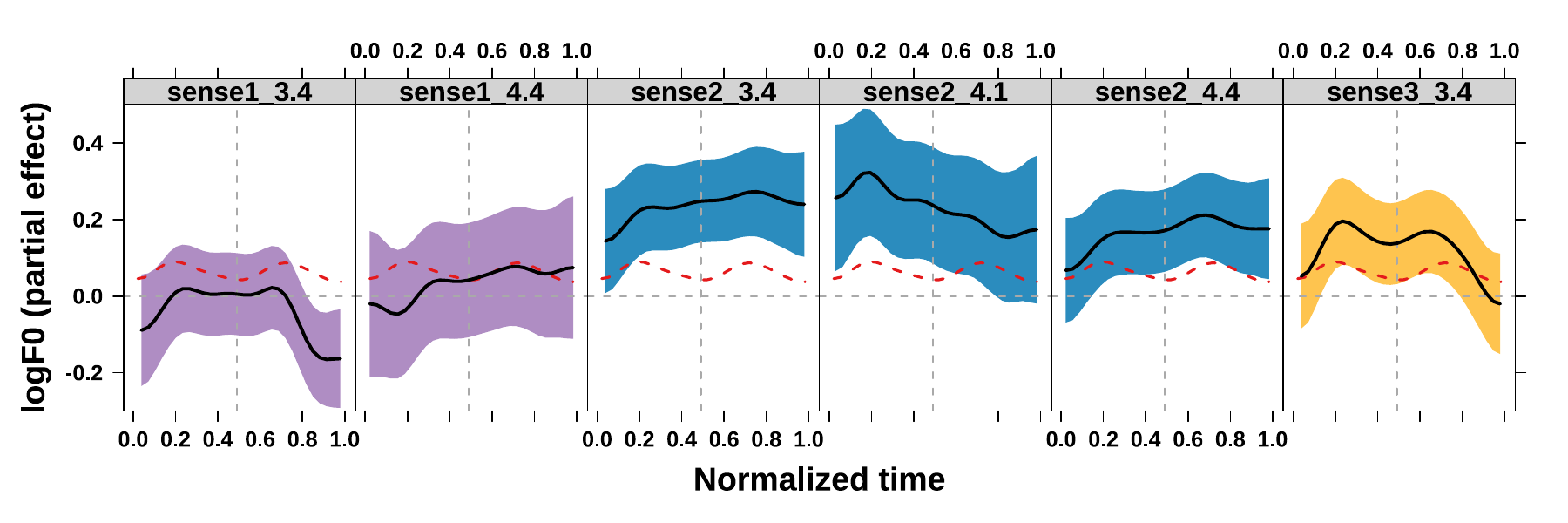}
  \caption{A selection of the predicted f0 contours for different \texttt{sense\_type} of 另外\ (\textit{ling4wai4}, `in addition') across tonal contexts. The predicted pitch contours represent the partial effect of the factor smooth for \texttt{sense\_type}, combined with the corresponding factor smooth for \texttt{tone\_pattern} (T4-T4 in this case). 
  The red dashed curves represent the partial effect of T4-T4 tone pattern alone, averaged across four tonal contexts, so the red dashed curve is the same across all panels. 
  Vertical dashed lines indicate the average syllable boundary, and the horizontal grey dashed line represents the y=0 reference line.}
  \label{fig:sense}
\end{figure}

Figure~\ref{fig:sense} displays the predicted tonal contours for different sense types of 另外 (\textit{ling4wai4}, 'in addition'), calculated by combining the partial effects of \texttt{sense\_type} and \texttt{tone\_pattern}. Similar to the red dashed curves in Figure~\ref{fig:word_pattern}, the red dashed curves in Figure~\ref{fig:sense} again represent the general tone pattern, which is T4-T4 in this case.
The three sense types of 另外 (\textit{ling4wai4}, `in addition') are: `others' (sense1), `totally different' (sense2), `in addition to' (sense3). 
The word 另外\ (\textit{ling4wai4}, ‘in addition’) exhibits clear variations across the three sense types compared to the general tone pattern. The pitch contours of sense 1 (shown in purple) are generally shifted below the general tone pattern, while those of sense 2 (shown in blue) are shifted above it. The pitch contour of sense 3 (shown in yellow) displays two rises, as in the general tone pattern, but is shifted upwards.


\subsection{Summary}

\noindent
This section addressed our first hypothesis, namely, that  the meanings of words co-determine the phonetic details of how the tones of these words are produced. Our results show that \texttt{word} emerged as a more powerful predictor than all other predictors.  Surprisingly, the variable importance of \texttt{word} was substantially greater than that of \texttt{tone\_pattern}. The strong effect of \texttt{word} that we observed is line with the results of \citet{jin2024corpus} and \citet{lu2024sandhi}. \citet{jin2024corpus} also observed, albeit for monosyllabic words, that \texttt{word} was a stronger predictor than \texttt{tone\_pattern}. In the study by \citet{lu2024sandhi}, however, the variable importance of \texttt{word} was similar to that of \texttt{tone\_pattern}.

A further analysis clarified that \texttt{sense\_type} is an even better predictor of pitch contours than \texttt{word}. The substantial improvement in model fit contributed by \texttt{sense\_type} provides further support for the hypothesis that it is words' meanings that  co-determine the fine detail of their pitch contours, replicating the findings of earlier studies \citep{chuang2024word, jin2024corpus, lu2024sandhi}.

\section{Theory-driven computational modeling} \label{sec:modeling}

\noindent
In this section, we turn to our second hypothesis, exploring whether the tonal realization of a given token can be predicted with reasonable accuracy based on its context-specific meaning using computational modeling. To do so, we make use of the general conceptual framework of the Discriminative Lexicon Model \citep[DLM][]{Heitmeier:Chuang:Baayen:2025}, a computationally implemented theory that was developed independently of the present data, but that turns out to provide exactly the right approach to predict tonal realization from semantics.

%

In the introduction, we already explained that the DLM seeks to predict words' forms from their meanings. Both forms and meanings are represented by numeric vectors, and in the simplest possible set-up, a linear mapping transforms a meaning vector into a form vector \citep[for mappings using deep neural networks, see][]{Heitmeier:Chuang:Baayen:2025}.  For the present study, we are not interested in predicting full word forms, but rather words' pitch contours.  What we need, then, are numerical representations of the present Mandarin word tokens' pitch contours on the one hand, and their meanings on the other hand. Following \citet{chuang2024word}, we represent words' forms using fixed-length vectors representing pitch contours, and we represent words' meanings using contextualized embeddings obtained with the GPT-2 transformer technology. Importantly, both the pitch vectors and the semantic vectors are context-specific, and thus vary from word token to word token. \citet{chuang2024word} demonstrated that the tonal contours of a given token with T2-T4 tone pattern can be predicted from its context-specific meaning with above-chance accuracy using a linear mapping.  In what follows, we consider whether this result generalizes to all 20 tone patterns attested for two-syllable words.  As a first step, we explain how we obtained fixed-length pitch vectors.

\subsection{Fixed-length pitch vectors}

\noindent
To implement a linear mapping within the DLM framework, given $n$ words, we need an $n \times p$ matrix $\bm{C}$ to represent words' pitch contours, and an $n \times q$ matrix $\bm{S}$ for words' meanings. Consider the form matrix $\bm{C}$, and recall that the tokens in our dataset have unequal numbers of pitch measurement points because tokens with longer durations contain more measurement points. Furthermore, the raw data include missing values due to gaps in the pitch contours. However, the row vectors of $\bm{C}$ need to have the same fixed length $p$. To achieve this, we used GAMs to obtain pitch contours represented by $p = 100$-dimensional vectors in normalized time.  There are several ways in which such fixed-length vectors can be generated, of which we explored three.


\begin{description}
    \item[Method I] The first method fitted separate GAMs to the f0 contours of each of the individual tokens, i.e., 4283 independent gam models, and then extracted the predicted contours. This method generates pitch contours that stay as close as possible to the empirical pitch measurements.  However, this method inevitably includes by-token measurement noise in the estimation of the contours.  In  the case of simple univariate linear regression, the predicted value for a data point (on the regression line) will deviate from the observed value for that data point; taking the observed data point as gold standard is at odds with statistical modeling. Similarly, for the present dataset of time-series of measurements, the observed curves are not the given gold standard. There are several sources of noise: articulatory stochastic noise in the articulation, noise in the audio recordings, and noise in the pitch measurements. Method I incorporates the combined noise originating from these sources. Therefore, Method I serves as a baseline that we expect to yield the least precise results.  Methods II and III implement two ways of reducing this by-observed pitch contour measure noise. 
    \item[Method II] The second method fitted a GAM to the f0 contours of all the tokens of words with a given tone pattern, extracting the smooth for time and the word-specific smooth, and combining these to obtain word-type-specific smooths. This method abstracts away from the influences of  contextual factors on the realization of pitch.  The resulting pitch vectors are identical for all the tokens of a given word type.  We anticipated that this would be the optimal situation for learning, as  within-type variation is eliminated.  This method also has a theoretical motivation, namely, that it is unlikely that the contextualized embeddings generated by an AI model will capture the full richness of the thought of  human speakers engaged in real, 30-minute long conversations. 
    \item[Method III] The third method, following \citet{chuang2024word}, obtains token-specific pitch vectors predicted by GAMs with all contextual factors included. For our data, we used the four GAMs fitted to the four tonal contexts, as reported above in Section ~\ref{sec:corpus}.  This method has the advantage, compared to Method I, of removing by-observation noise. Furthermore, compared to Method II, it has the advantage of having pitch vectors that vary from token to token.  Thus, this method is optimal for detecting the extent to which by-token semantics and by-token phonetics are aligned. The more the contextualized embeddings diverge from the true semantic intentions of the speakers, the less well this method will perform. 
\end{description}


\noindent
After obtaining the estimated pitch vectors from the GAMs, we applied by-token normalization by centering and scaling each predicted pitch vector. By doing so, the mapping from meaning to form is forced to learn to predict the shape of pitch contours rather than the absolute pitch values of each token, which vary substantially across word types and speakers.

\subsection{Contextualized embeddings}

\noindent
We made use of Contextualized Embeddings (CEs) to represent words' meanings. Word embeddings (semantic vectors) represent meanings in a distributed manner, building on the hypothesis that similar words occur in similar contexts \citep{harris1954distributional,Landauer:Dumais:1997,mikolov2013distributed}. The first generation of semantic embeddings, such as \texttt{fastText} \citep{bojanowski2017enriching}, is fully determined by words' orthographic forms. However, a single orthographic form can express different meanings (e.g., English `bank'), or different senses (e.g., Mandarin 水平\ (\textit{shui3ping2}, `level or horizontal position' or `skill or proficiency')). Typically, the context in which a word is used provides disambiguating information.  Contextualized Embeddings (CEs) were developed to provide token-specific, context-sensitive embeddings that capture  the subtle differences in what a word may actually mean in context. 

The CEs used in the current study were derived from a pre-trained unidirectional language model based on the GPT-2 architecture, developed by CKIP, Academia Sinica. Each token in our dataset was assigned a 768-dimensional vector representing its contextualized embedding. 

\begin{figure} [htbp]
    \centering
    \includegraphics[width=\textwidth]{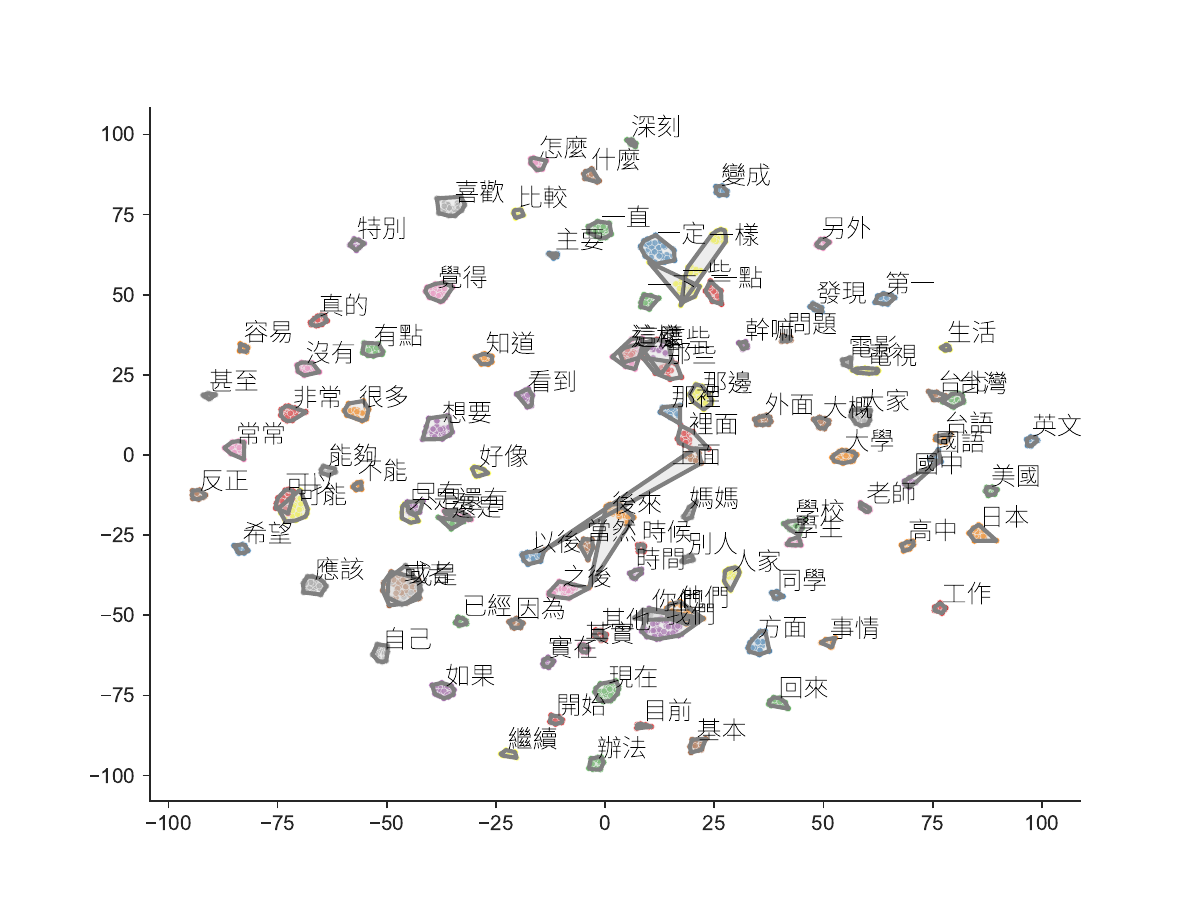}
    \caption{Contextualized embeddings, obtained from a pre-trained Chinese GPT-2 model, are shown in a two-dimensional plane obtained with t-SNE. Convex hulls (grey polygons) highlight the clusters of word types.}
    \label{fig:cluster}
\end{figure}

To inspect the quality of the contextualized embedding space,  we reduced the 768-dimensional semantic space to two dimensions using t-SNE \citep{van2008visualizing}. Figure~\ref{fig:cluster} displays embeddings in the resulting 2-D plane, with convex hulls highlighting clusters of tokens corresponding to different word types. Tokens clearly cluster by word, as expected. Furthermore, some semantically related words have clusters that are close to each other. For instance, in the  middle-right of the Figure, the tokens of 大學\ (\textit{da4xue2}, `university'), 學校\  (\textit{xue2xiao4}, `school'), 國中\ (\textit{guo2zhong1}, `middle school' ), and 高中\ (\textit{gao1zhong1}, `high school') occur closely together, which makes sense as they are all semantically related to educational institutions. Other school-related words such as 學生\ (\textit{xue2sheng1}, 'students') and 老師\ (\textit{lao3shi1}, 'teacher') also appear near these words.  Some word clusters contain outliers. For instance, in the center of the figure, 上面\ (\textit{shang4mian4}, 'above') has an outlier positioned near 以後\ (\textit{yi3hou4}, 'in the future'), and 後來\ (\textit{hou4lai2}, 'afterwards') has an outlier near 之後\ (\textit{zhi1hou4}, 'after').


\subsection{Method}

Modeling was conducted using the same dataset that we used above for the word-type-based analysis (see Table ~\ref{tab:dataset}), which contains 4,283 tokens. This dataset, comprising all four tonal contexts, was split into a training dataset (80.39\%) and a testing dataset (19.61\%). Every word type was represented in both the training and testing data, with tokens per word being split roughly proportionally with 80\% in the training dataset and 20\% in the testing dataset.

Using the training data, we computed a linear mapping $\bm{G}$ from a $3443 \times 768$ semantic matrix $\bm{S}$ to a $3443 \times 100$ form matrix $\bm{C}$ by solving $\bm{S}\bm{G}=\bm{C}$ \citep[for technical details, see][]{gahl2024time,Heitmeier:Chuang:Baayen:2025}. We then evaluated the quality of the mapping for both the training and the testing dataset. 

The accuracy of a predicted pitch vector $\hat{\bm{c}}$ was evaluated as follows. For a given $\hat{\bm{c}}$, we calculated its Euclidean distance to all gold-standard pitch vectors in $\bm{C}$. We then identified its closest form neighbor of $\hat{\bm{c}}$. If this nearest neighbor was a token of the same word type as the target token, the predicted form vector was assessed as correct; otherwise, it was considered incorrect.

\subsection{Results}

\noindent
We estimated three linear mappings from the same semantic matrix $\bm{S}$ with CEs to three different form matrices $\bm{C}$, one for each of the three kinds of smoothed pitch contours introduced above. 

The mean accuracy of method \uppercase\expandafter{\romannumeral1}  was 2.8\% on the training dataset and 1.4\% on the testing dataset. The mean accuracy of method \uppercase\expandafter{\romannumeral2} was 23.5\% on the training dataset and 15.1\% on the testing dataset. The mean accuracy of method \uppercase\expandafter{\romannumeral3}  was 12.3\% on the training dataset and 7.7\% on the testing dataset. All accuracies were above a permutation baseline of 0.4\% and a majority baseline 1.3\%, albeit by only a small margin for method \uppercase\expandafter{\romannumeral1}.  That method \uppercase\expandafter{\romannumeral1} has the lowest accuracy is unsurprising, fitting GAMs to individual pitch contours unavoidably comes with overfitting and a loss of generalizability.  Methods \uppercase\expandafter{\romannumeral2} and \uppercase\expandafter{\romannumeral3} gain strength from other tokens and incorporate less by-item observation noise.  For these two methods, the mapping from meaning to pitch contours is substantially more accurate than would be expected under chance conditions.  The best-performing method is method \uppercase\expandafter{\romannumeral2}, which abstracts away from the influences of contextual factors on the realization of pitch.  This suggests that some abstraction away from the immediate context is helpful, possibly because the contextualized embeddings are not precise enough. After all, these embeddings come from a general large language model trained on large volumes of data that most likely diverge considerably for the language experience of the speakers interviewed for the Corpus of Spoken Taiwan Mandarin. 

\begin{figure}[htbp]
  \centering
  \includegraphics[width=\textwidth]{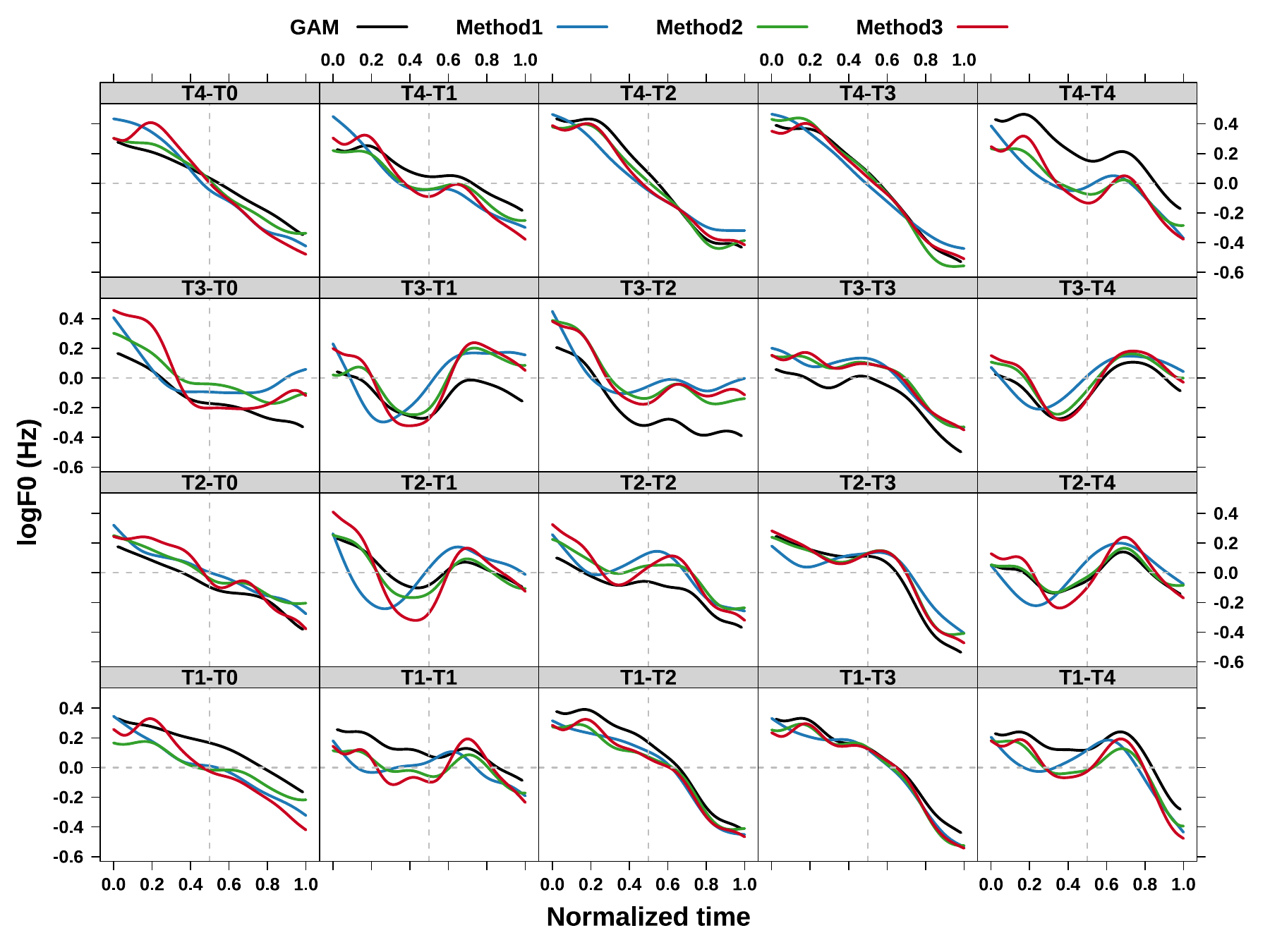}
  \caption{Pitch contours of 20 tone patterns in selected four tonal contexts. The black curves present  pitch contours identified by GAM, estimated by the partial effect for \texttt{tone\_pattern}, combined with a general contours of time female speakers (shown as the red curves in Figure~\ref{fig:pattern}, which is reproduced here).  The blue, green, and red curves represent pitch contours predicted from the centroid of the contextualized embeddings using the three methods, respectively. For the blue curves, form vectors were obtained with GAM smooths fitted to individual word tokens. For the green curves, form vectors were obtained from a GAM fitted to the tokens of all words with a given tone pattern. For the red curves, form vectors were obtained with GAM smooths that included all contextual factors.}
  \label{fig:GAM_LDL}
\end{figure}

The results obtained with especially methods \uppercase\expandafter{\romannumeral2} and \uppercase\expandafter{\romannumeral3} clarify that there appears to be considerable isomorphy between the contextualized embedding space and the pitch space of word tokens. This isomorphy implies that if we take the most typical embedding for a given tone pattern and map it into the pitch space, the resulting predicted pitch contours should closely resemble the pitch contours identified by the GAM for that tone pattern.   Figure~\ref{fig:GAM_LDL} shows that this prediction is on the right track.  The pitch contours shown in black are the contours predicted by the GAMs for the different tone patterns.  They represent the best de-noised estimates of the average tone-pattern-specific pitch contours, and serve as our gold-standard pitch contours.  These GAM-based pitch contours were obtained by first extracting the tone-pattern specific pitch contours for each of the four tonal contexts, and then averaging these. (These GAM-based contours were shown above in red in Figure~\ref{fig:pattern} before). An alternative method, that results in nearly indistinguishable pitch contours, combines the data for all four contexts, and adds the tone-pattern specific 
contours to the general contour for female speakers. 

We now consider how well these average pitch contours can be predicted from words' contextualized embeddings. The most typical embedding for a given tone pattern can be approximated by calculating the centroid of the contextualized embeddings of the tokens with this particular tone pattern. The centroid is simply the mean of the semantic vectors. When we think of embeddings as points in a high-dimensional space, the centroid is located at the center of these points. To obtain the centroid of a given tone pattern, we first obtained the centroid of every relevant word type by averaging the CEs of its tokens. We then obtained the centroid of the tone pattern by averaging the centroids of the word types associated with that tone pattern. In this way, every word type is given equal weight when determining the centroids for the tone patterns.

In order to get a sense of the semantics represented by these centroid vectors, we calculated, for each tone pattern, which contextualized embeddings are closest to the corresponding centroids. Table~\ref{tab:top2} lists, for each tone pattern, the two word types with embeddings that are closest to the centroids.  These two word types provide an indication of the prototypical meaning of a tone pattern. For example, 她們\ (\textit{ta1men0}, `they, female') and 他們\ (\textit{ta1men0}, `they, male') are the most prototypical word types for T1-T0 tone pattern.  The tone pattern T2-T0 appears to have 兒子 (\textit{er2zi0}, `son') and 孩子\ (\textit{hai2zi0}, `child') as prototypical members.  Most tone patterns, however, are characterized by function words.

\begin{table}[htbp]
\caption{The top two word types which have contextualized embeddings that are closest to the centroid embedding of the 20 tone patterns. Proximity is evaluated using Euclidean distance. }
\centering
\begin{tabular}{lll}
  \hline
 \textbf{Tone pattern} & \textbf{Top one closest word type} & \textbf{Top two closest word type} \\ 
  \hline
 T1-T0  & 她們\ (\textit{ta1men0}, `they') & 他們\ (\textit{ta1men0}, `they') \\ 
 T1-T1  & 一些\ (\textit{yi1xie1}, `some') & 一般\ (\textit{yi1ban1}, `general') \\ 
 T1-T2  & 當然\ (\textit{dang1ran2}, `of course') & 之前\ (\textit{zhi1qian2}, `before') \\ 
 T1-T3  & 剛好\ (\textit{gang1hao3}, `just right') & 一起\ (\textit{yi1qi3}, `together') \\ 
 T1-T4  & 之後\ (\textit{zhi1hou4}, `afterwards') & 之類\ (\textit{zhi1lei4}, `and so on') \\ 
 T2-T0  & 兒子\ (\textit{er2zi0}, `son') & 孩子\ (\textit{hai2zi0}, `child') \\ 
 T2-T1  & 人家\ (\textit{ren2jia1}, `others') & 國中\ (\textit{guo2zhong1}, `middle school') \\ 
 T2-T2  & 其實\ (\textit{qi2shi2}, `actually') & 別人\ (\textit{bie2ren0}, `others') \\ 
 T2-T3  & 還有\ (\textit{hai2you3}, `also') & 還好\ (\textit{hai2hao3}, `it's okay') \\ 
 T2-T4  & 然後\ (\textit{ran2hou4}, `then') & 一樣\ (\textit{yi2yang4}, `the same') \\ 
 T3-T0  & 你們\ (\textit{ni3men0}, `you all') & 我們\ (\textit{wo3men0}, `we') \\ 
 T3-T1  & 很多\ (\textit{hen3duo1}, `many') & 女生\ (\textit{nv3sheng1}, `girls') \\ 
 T3-T2  & 起來\ (\textit{qi3lai2}, `get up') & 以前\ (\textit{yi3qian2}, `before') \\ 
 T3-T3  & 只有\ (\textit{zhi3you3}, `only') & 有點\ (\textit{you3dian3}, `a bit') \\ 
 T3-T4  & 以後\ (\textit{yi3hou4}, `afterwards') & 好像\ (\textit{hao3xiang4}, `seems like') \\ 
 T4-T0  & 這麼\ (\textit{zhe4me0}, `so') & 那麼\ (\textit{na4me0}, `that') \\ 
 T4-T1  & 那些\ (\textit{na4xie1}, `those') & 那邊\ (\textit{na4bian1}, `over there') \\ 
 T4-T2  & 個人\ (\textit{ge4ren2}, `individual') & 不然\ (\textit{bu4ran2}, `otherwise') \\ 
 T4-T3  & 到底\ (\textit{dao4di3}, `after all') & 那裡\ (\textit{na4li3}, `there') \\ 
 T4-T4  & 算是\ (\textit{suan4shi4}, `considered as') & 上面\ (\textit{shang4mian4}, `above') \\ 
  \hline
\end{tabular}
\label{tab:top2}
\end{table}

To obtain the pitch contours predicted for the tone patterns, we provided the centroids of the 20 tone patterns as input to the three linear mappings defined above. The resulting predicted pitch contours are shown in Figure~\ref{fig:GAM_LDL}.  Each panel in this trellis graph presents the estimates for a given tone pattern. The gold-standard pitch contours (obtained with our GAM models as described above) are presented in black, and the contours predicted by the three DLM mappings are color-coded. The contours obtained with method \uppercase\expandafter{\romannumeral1} are shown in blue, those obtained with method \uppercase\expandafter{\romannumeral2} in green, and those obtained with method \uppercase\expandafter{\romannumeral3} in red. For all three methods, the resulting predicted contours are similar, and often remarkably similar, to the gold-standard contours.

\begin{figure} [htbp]
    \centering
    \includegraphics[width=0.8\textwidth]{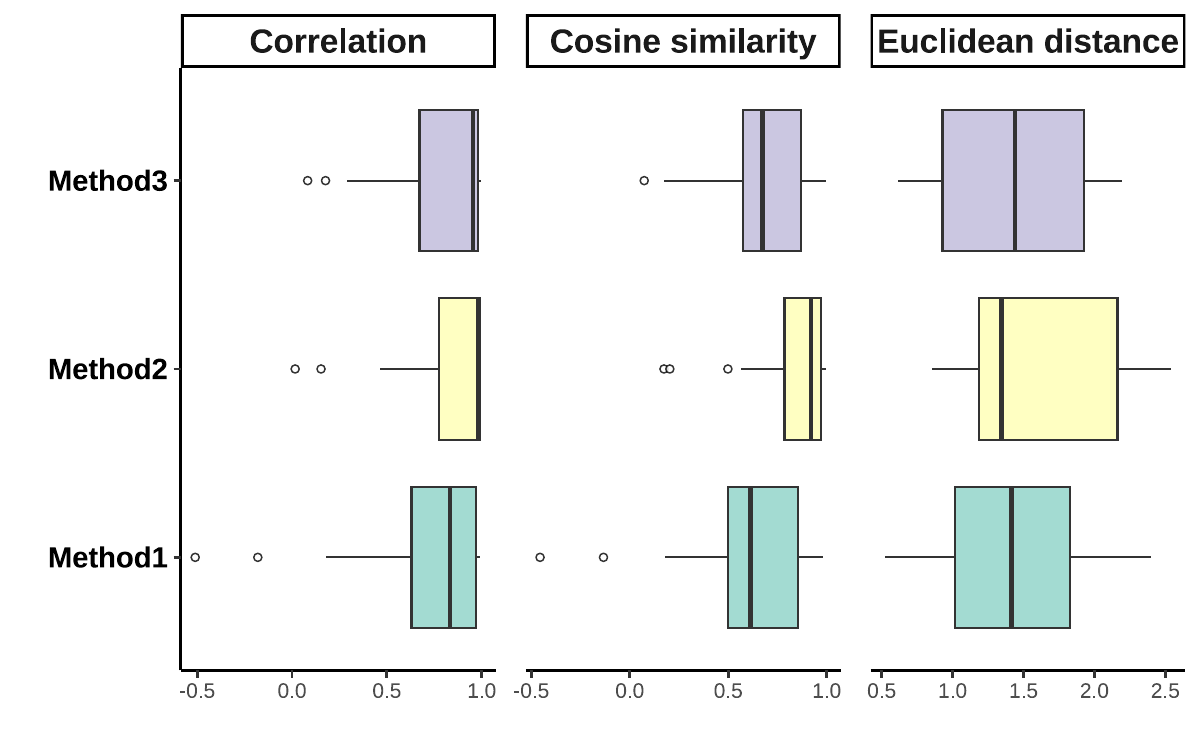}
    \caption{Boxplot showing the correlation, cosine similarity, and Euclidean distance between GAM-predicted pitch contours and DLM-derived pitch contours across 20 tone patterns. }
    \label{fig:similarity}
\end{figure}

To assess the similarity between the gold-standard pitch contours and the pitch contours predicted using the meaning-to-pitch mappings, we first calculated the cosine similarity, averaged across the 20 tone patterns, between the gold-standard contours and each of the three DLM pitch contours. The contours from method \uppercase\expandafter{\romannumeral2} show a closer fit (cosine similarity 0.81) compared to the contours from method \uppercase\expandafter{\romannumeral1} (cosine similarity 0.59) and \uppercase\expandafter{\romannumeral3} (cosine similarity 0.66). The mean correlation between GAM-predicted pitch contours and three DLM-derived contours is 0.69, 0.82, and 0.78, respectively. However, when using Euclidean distance to evaluate similarity, method \uppercase\expandafter{\romannumeral2} scores slightly worse than the other two (1.48, 1.57, and 1.43, respectively). Figure~\ref{fig:similarity} presents the distributions of these three measures for the three methods, using boxplots. 
Regardless of how exactly the precision of the predictions is evaluated at the level of centroids, the three methods appear to perform with comparable accuracy, with a slight advantage for Method \uppercase\expandafter{\romannumeral2} when evaluated with the correlation and cosine similarity measures.

\subsection{Summary}


\noindent
In this section, we have shown that a simple linear mapping can predict the realization of token-specific pitch contours from its token-specific meaning in context with above-chance accuracy. This finding extends the earlier results of \citet{chuang2024word}, which focused on disyllabic words with one specific tone pattern only (the rise-fall tone pattern T2-T4).  Mapping accuracy for all 20 tone patterns is unsurprisingly somewhat lower that that observed by \citet{chuang2024word} for the T2-T4 tone pattern (30\%–40\% for training data and 27\%–35\% for testing data). Nevertheless, even for the present more varied dataset, accuracy is substantially above the majority baseline.  This result is surprising in the light of the measurement noise that is inevitably present in both our pitch measurements and in the contextualized embeddings.  The contextualized embeddings represent the knowledge of an artificial intelligence, trained on vast amounts of texts. The embeddings it generates must diverge from the meanings that the individual speakers had in mind. Nevertheless, the contextualized embeddings are sufficiently precise to enable far above chance prediction accuracy for word tokens' pitch contours.  Interestingly, the 20 canonical tone patterns are surprisingly well approximated by projecting the centroids of the contextualized embeddings of the words with these tone patterns into the f0 space.  In other words, the average pitch contours identified by the GAMs correspond to average contextualized embeddings in semantic space.


\section{General discussion} \label{sec:gendisc}

\noindent
The current study investigated the realization of pitch contours of disyllabic words in a corpus of spontaneous spoken Taiwan Mandarin. We first made use of the Generalized Additive Models to decompose f0 contours into a series of nonlinear functions of normalized time, with each function representing the way in which a predictor modulates the pitch contour over time. A range of predictors was taken into account, including normalized time, gender, tonal context, tone pattern, speech rate, word position, bigram probability, and speaker.  Surprisingly, the GAMs provided strong support for word-specific modulations of the pitch contours. Replacing word by word sense further improved model fit, which suggests that the effect of word may be semantic in nature.  If so, the theory of the Discriminative Lexicon Model predicts that it should be possible to well approximate the token-specific pitch contours observed in the corpus with predicted token-specific pitch contours obtained with mappings taking the  contextualized embeddings of the words in the corpus as input, and producing the corresponding pitch contours as output.  We found that indeed a mapping from GPT-2 generated contextualized embeddings to 100-dimensional fixed-length pitch vectors predicts words' pitch contours with accuracies that are far above a majority choice baseline. Thus, our study successfully extends the meaning-to-pitch mapping from the T2-T4 tone pattern studied by  \citep{chuang2024word} to all tone patterns in Taiwan Mandarin. Our study also dovetails well with the evidence for the importance of word and sense as co-determinants of pitch contours reported by \citet{lu2024sandhi} and \citet{jin2024corpus}.


A remarkable finding is that words and their meanings co-determine the realization of the f0 contours in disyllabic words with effect sizes that considerably exceed those of tone pattern. This finding for disyllabic words aligns with previous research on Mandarin monosyllabic words \citep{jin2024corpus}, which reported that while the effect of tone pattern on pitch contours is modest, the effect of word is substantial. For disyllabic words, the stronger effect of word largely overshadows the effect of tone pattern. 

\begin{figure}[htbp]
  \centering
  \includegraphics[width=\textwidth]{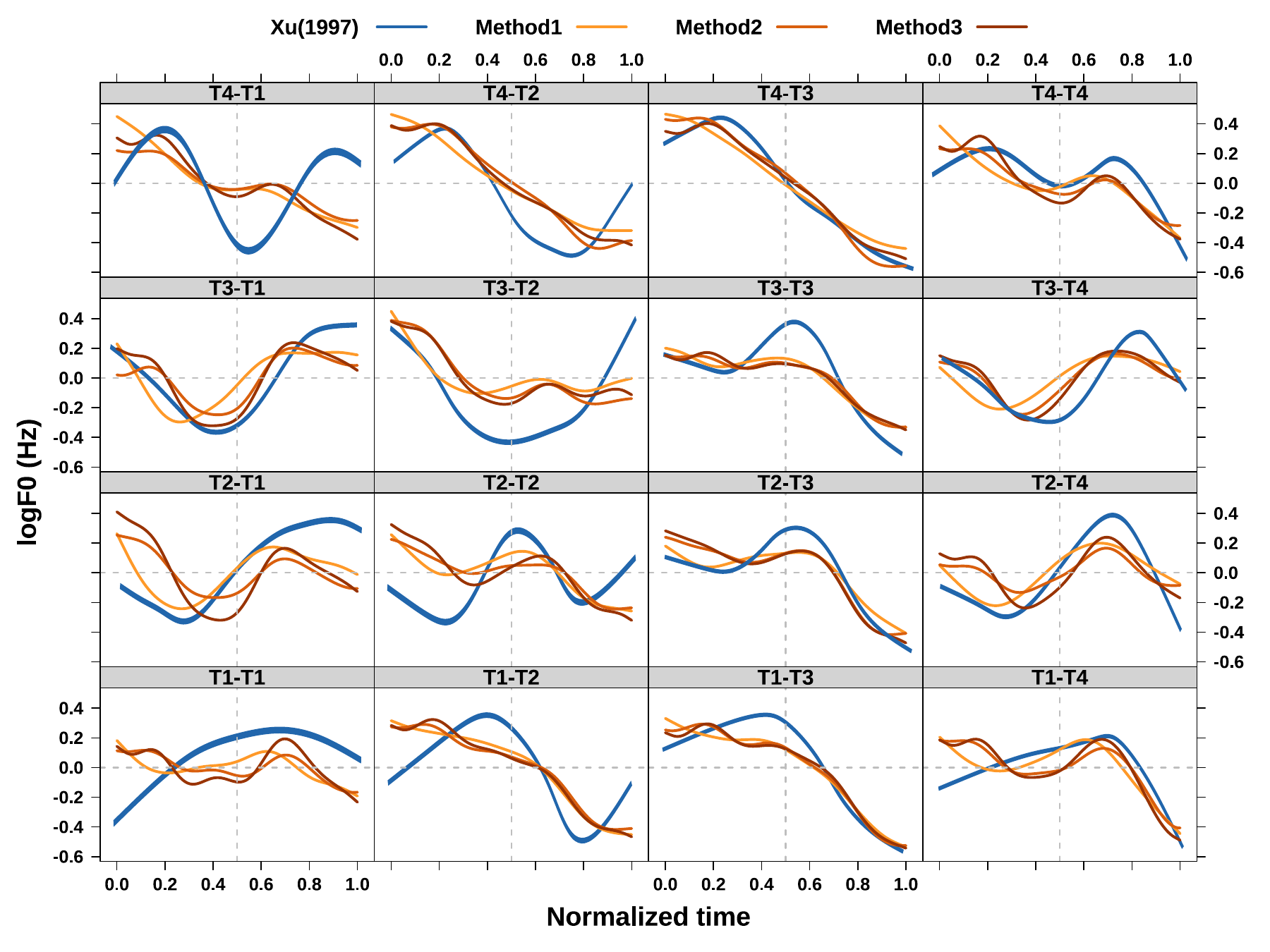}
  \caption{The f0 contours for 16 tone patterns from the current study, based on the Corpus of Spoken Taiwan Mandarin, are compared with f0 contours from a previous study \citep{xu1997contextual} on carefully controlled laboratory speech. The blue curves represent the f0 contours for 16 tone patterns from the controlled laboratory speech, reproduced from Figure 3 in \citet{xu1997contextual}. The orange curves correspond to the three LDL-predicted pitch contours from Method \uppercase\expandafter{\romannumeral1}, \uppercase\expandafter{\romannumeral2}, and \uppercase\expandafter{\romannumeral3}, as shown in Figure~\ref{fig:GAM_LDL}, and are reproduced here. These f0 contours are overlaid for comparison. Since the neutral tone (T0) was not included in \citet{xu1997contextual}, only 16 bi-tonal combinations are presented here.}
  \label{fig:xu1997}
\end{figure}

Our results suggest that there are not only  remarkable similarities, but also some clear differences, between tonal realization in laboratory speech and tonal realization in the Corpus of Spoken Taiwan Mandarin. \citet{xu1997contextual} analyzed the pitch contours of 16 bi-tonal combinations using the /ma-ma/ sequence. Among these combinations, only \textit{ma1ma1} corresponds to a real word in Mandarin, 妈妈 (\textit{ma1ma1}, ``mother''); all the others are nonsensical combinations that are unnatural for native speakers. In their study, the f0 contours were carefully controlled, accounting for factors such as gender and speaking rate.  Although laboratory speech and spontaneous speech differ in many ways, it is still informative to compare the two registers.  We therefore reproduced Figure 3 from \citet{xu1997contextual} (blue curves) and overlaid it with the DLM-derived f0 contours from Figure~\ref{fig:GAM_LDL}  (orange curves). In Figure~\ref{fig:xu1997}, the pitch contours from \citet{xu1997contextual} and our predicted contours are remarkably similar for several tone patterns, including T4-T4, T2-T3, and T2-T4. However, some tone patterns, such as T1-T1 and T1-T2, exhibit noticeable differences in pitch contours. These differences can be attributed to dialect differences, differences between spontaneous speech and laboratory speech, and differences between meaningful and meaningless words. 

At this point, it might be objected that in this approach to Mandarin tone, it is unclear how tone sandhi could be accounted for. How would it be possible that, if indeed every word has its own pitch contour, all words with the T3-T3 tone pattern undergo the same phonological process, such that they become indistinguishable from words with the T2-T3 tone pattern?  Our answer to this question has an empirical part and a theoretical part. 
  
On the empirical side, in conversational Taiwan Mandarin, the two tone patterns are basically identical. For instance, in Figure~\ref{fig:GAM_LDL}, the tone patterns for T2-T3 and T3-T3 are quite similar. A detailed study of this tone sandhi \citep{lu2024sandhi} supports complete neutralization for Taiwan Mandarin.  In other words, the words with the T3-T3 tone pattern can simply be re-classified as words with the T2-T3 tone pattern.  There is no need to call on a rule of tone sandhi.   In fact, even for standard Mandarin, as gauged by \citet{xu1997contextual}, the differences between T3-T3 and T2-T3 are hardly visible to the eye. However, T3-T3 tone sandhi has been reported to be incomplete for standard Mandarin \citep{yuan20143rd}.


This brings us to the theoretical aspect of the question raised above, namely, how to account for tone sandhi processes in general.  Within the framework of the Discriminative Lexicon model, as a model for highly automatized lexical processing, it is impossible to derive forms from forms, a method widely used in educational settings. Forms are predicted from meanings. Importantly, Figures~\ref{fig:GAM_LDL} and ~\ref{fig:xu1997} show that the tone contours associated with tone patterns emerge straightforwardly from the meaning-to-form mapping, without the model ever being informed about tone patterns. In other words, a `word and paradigm' approach \citep{Blevins:2016} to tonal realization appears to be quite feasible.

A question for further research is how to interpret the present findings for tone patterns with the neutral tone. As can be seen in Figure~\ref{fig:GAM_LDL}, for three of the four tone patterns, the overall pitch contour appears to be an almost linearly descending pitch contour. This could be viewed as another instance of tone sandhi in classical phonology, whereas within the DLM, this patterning would follow from words' contextual meanings. We leave this question for further research.


The results obtained in the present study have several theoretical implications. First, we have documented that the mapping from context-sensitive meaning to pitch contours is machine-learnable.  It remains an open question whether human learners also generate pitch contours from semantics.  The finding that just a linear mapping (from a statistical prespective, a straightforward multivariate multiple regression model) is all that is needed suggests that human speakers should also be able to learn this simple mapping between meaning and form.  Importantly, our results are based on patterns of usage in a corpus of unscripted spontaneous speech, and the mere existence of these patterns indicates that language users must be absorbing community norms for tonal realization, albeit most likely subliminally.  

Second, our findings challenge the axioms of the arbitrariness of the sign and the dual articulation of language.  If the relation between form and meaning would be truly and fundamentally arbitrary, this would imply learning words and their meanings is extremely difficult, and would not allow any generalization. All that can be done is learn by heart that a form $x$ is associated with a meaning $y$.  However, our simple linear mapping generalizes to held-out data. This falsifies the axiom that the relation between form and meaning (here, pitch and meaning) is completely arbitrary.

Third, preliminary results, reported in \citet{chuangwords}, suggest that English two-syllable words with left stress also have pitch contours that have strong word-specific pitch components.  The study by \citet{Schmitz2025} reports similar findings for English three-constituent compounds.  The accumulating evidence poses a new challenge for linguistics: understanding why these isomorphies between form and meaning exist, irrespective of whether a language is a tone language or a stress language.


\section*{Funding}
This work was supported by the European Research Council under Grant SUBLIMINAL (\#101054902) awarded to R. Harald Baayen.

\section*{Acknowledgements}
The authors thank Yu-Hsiang Tseng for identifying word sense for the dataset in the current manuscript.

\newpage
\appendix

\section*{Appendix 1: dataset and model summary} \label{sec:appendix1}
\bigbreak

\renewcommand{\thetable}{A.\arabic{table}}

\setcounter{table}{0}

\begin{table}[htbp]
\caption{Overview of the four sub-datasets grouped by the four tonal contexts, with the number of word types for each tone pattern.}
\centering
\begin{tabular}{lcccc}
  \toprule
  \textbf{Tone Pattern} & \multicolumn{4}{c}{\textbf{Tonal Context}} \\ 
  \cmidrule(lr){2-5}
  & \textbf{4.4} & \textbf{3.4} & \textbf{4.1} & \textbf{4.0} \\ 
  \midrule
  T1-T0  & 4  & 4  & 4  & 2  \\ 
  T1-T1  & 15 & 15 & 12 & 12 \\ 
  T1-T2  & 18 & 15 & 17 & 11 \\ 
  T1-T3  & 8  & 8  & 6  & 6  \\ 
  T1-T4  & 23 & 19 & 18 & 22 \\ 
  T2-T0  & 7  & 6  & 6  & 4  \\ 
  T2-T1  & 11 & 9  & 9  & 11 \\ 
  T2-T2  & 14 & 9  & 11 & 8  \\ 
  T2-T3  & 14 & 12 & 13 & 12 \\ 
  T2-T4  & 23 & 19 & 22 & 12 \\ 
  T3-T0  & 4  & 3  & 4  & 4  \\ 
  T3-T1  & 10 & 8  & 6  & 7  \\ 
  T3-T2  & 16 & 12 & 11 & 12 \\ 
  T3-T3  & 11 & 10 & 9  & 8  \\ 
  T3-T4  & 18 & 15 & 17 & 11 \\ 
  T4-T0  & 3  & 4  & 3  & 1  \\ 
  T4-T1  & 10 & 9  & 12 & 11 \\ 
  T4-T2  & 24 & 20 & 17 & 15 \\ 
  T4-T3  & 9  & 8  & 8  & 5  \\ 
  T4-T4  & 46 & 35 & 45 & 36 \\ 
  \midrule
  \textbf{Total} & \textbf{288} & \textbf{240} & \textbf{250} & \textbf{210} \\ 
  \bottomrule
\end{tabular}
\label{tab:counts}
\end{table}

\begin{table}[htbp]
\caption{Summary of the model fitted with \texttt{word} for the \texttt{4.4} tonal context dataset} 
\centering
\begin{tabular}{lrrrr}
   \hline
A. parametric coefficients & Estimate & Std. Error & t-value & p-value \\ 
  (Intercept) & 5.2965 & 0.0251 & 210.6574 & $<$ 0.0001 \\ 
  gendermale & -0.5283 & 0.0343 & -15.3930 & $<$ 0.0001 \\ 
   \hline
B. smooth terms & edf & Ref.df & F-value & p-value \\ 
  s(normalized\_t):genderfemale & 1.0020 & 1.0025 & 24.4870 & $<$ 0.0001 \\ 
  s(normalized\_t):gendermale & 2.6783 & 2.9183 & 7.6161 & $<$ 0.0001 \\ 
  s(speaker) & 51.0328 & 53.0000 & 72.0201 & $<$ 0.0001 \\ 
  s(normalized\_t,word) & 1832.2415 & 2592.0000 & 6.9980 & $<$ 0.0001 \\ 
  s(normalized\_t,tone\_pattern) & 111.9037 & 179.0000 & 2.3357 & $<$ 0.0001 \\ 
  s(speech\_rate):genderfemale & 2.1305 & 2.5343 & 3.8114 & 0.0126 \\ 
  s(speech\_rate):gendermale & 2.3418 & 2.7025 & 9.7111 & $<$ 0.0001 \\ 
  ti(normalized\_t,speech\_rate) & 2.9792 & 3.0132 & 30.9735 & $<$ 0.0001 \\ 
  s(norm\_utt\_pos) & 1.7416 & 2.1134 & 101.1452 & $<$ 0.0001 \\ 
  ti(normalized\_t,norm\_utt\_pos) & 6.6304 & 7.9364 & 6.7134 & $<$ 0.0001 \\ 
  s(bg\_prob\_prev) & 2.8803 & 2.9825 & 32.2276 & $<$ 0.0001 \\ 
  ti(normalized\_t,bg\_prob\_prev) & 4.9995 & 6.2443 & 2.3813 & 0.0252 \\ 
  s(bg\_prob\_fol) & 1.0072 & 1.0139 & 4.8743 & 0.0265 \\ 
  ti(normalized\_t,bg\_prob\_fol) & 7.8925 & 8.6541 & 9.2356 & $<$ 0.0001 \\ 
   \hline
\end{tabular}
\label{tab:model1}
\end{table}

\begin{table}[htbp]
\caption{Summary of the model fitted with \texttt{word} for the \texttt{3.4} tonal context dataset} 
\centering
\begin{tabular}{lrrrr}
   \hline
A. parametric coefficients & Estimate & Std. Error & t-value & p-value \\ 
  (Intercept) & 5.3009 & 0.0239 & 222.0659 & $<$ 0.0001 \\ 
  gendermale & -0.5151 & 0.0309 & -16.6894 & $<$ 0.0001 \\ 
   \hline
B. smooth terms & edf & Ref.df & F-value & p-value \\ 
  s(normalized\_t):genderfemale & 2.8018 & 2.9407 & 11.3801 & $<$ 0.0001 \\ 
  s(normalized\_t):gendermale & 1.0026 & 1.0035 & 0.6147 & 0.4328 \\ 
  s(speaker) & 49.9064 & 54.0000 & 24.2216 & $<$ 0.0001 \\ 
  s(normalized\_t,word) & 1430.9843 & 2160.0000 & 5.8479 & $<$ 0.0001 \\ 
  s(normalized\_t,tone\_pattern) & 92.8198 & 179.0000 & 1.3530 & $<$ 0.0001 \\ 
  s(speech\_rate):genderfemale & 2.3721 & 2.7269 & 9.3361 & $<$ 0.0001 \\ 
  s(speech\_rate):gendermale & 2.1719 & 2.5746 & 3.4047 & 0.0351 \\ 
  ti(normalized\_t,speech\_rate) & 6.7124 & 7.9532 & 4.9905 & $<$ 0.0001 \\ 
  s(norm\_utt\_pos) & 1.0004 & 1.0009 & 144.2261 & $<$ 0.0001 \\ 
  ti(normalized\_t,norm\_utt\_pos) & 6.4372 & 7.6506 & 4.8472 & $<$ 0.0001 \\ 
  s(bg\_prob\_prev) & 2.4885 & 2.7657 & 15.9136 & $<$ 0.0001 \\ 
  ti(normalized\_t,bg\_prob\_prev) & 8.5179 & 8.8649 & 26.3242 & $<$ 0.0001 \\ 
  s(bg\_prob\_fol) & 1.0023 & 1.0044 & 8.8528 & 0.0029 \\ 
  ti(normalized\_t,bg\_prob\_fol) & 2.3852 & 2.7236 & 3.6728 & 0.0093 \\ 
   \hline
\end{tabular}
\label{tab:model2}
\end{table}

\begin{table}[htbp]
\caption{Summary of the model fitted with \texttt{word} for the \texttt{4.1} tonal context dataset} 
\centering
\begin{tabular}{lrrrr}
   \hline
A. parametric coefficients & Estimate & Std. Error & t-value & p-value \\ 
  (Intercept) & 5.2709 & 0.0251 & 210.1867 & $<$ 0.0001 \\ 
  gendermale & -0.4851 & 0.0331 & -14.6522 & $<$ 0.0001 \\ 
   \hline
B. smooth terms & edf & Ref.df & F-value & p-value \\ 
  s(normalized\_t):genderfemale & 1.0003 & 1.0004 & 14.3424 & 0.0002 \\ 
  s(normalized\_t):gendermale & 2.4742 & 2.7866 & 3.8965 & 0.0197 \\ 
  s(speaker) & 50.3703 & 54.0000 & 28.1287 & $<$ 0.0001 \\ 
  s(normalized\_t,word) & 1520.9285 & 2250.0000 & 5.5411 & $<$ 0.0001 \\ 
  s(normalized\_t,tone\_pattern) & 88.4589 & 179.0000 & 1.2859 & $<$ 0.0001 \\ 
  s(speech\_rate):genderfemale & 1.0005 & 1.0010 & 5.3846 & 0.0203 \\ 
  s(speech\_rate):gendermale & 1.9702 & 2.3439 & 2.9261 & 0.0393 \\ 
  ti(normalized\_t,speech\_rate) & 6.2226 & 7.4529 & 14.0239 & $<$ 0.0001 \\ 
  s(norm\_utt\_pos) & 1.0006 & 1.0011 & 28.5619 & $<$ 0.0001 \\ 
  ti(normalized\_t,norm\_utt\_pos) & 7.3974 & 8.3940 & 2.4661 & 0.0054 \\ 
  s(bg\_prob\_prev) & 2.0687 & 2.4374 & 12.4622 & $<$ 0.0001 \\ 
  ti(normalized\_t,bg\_prob\_prev) & 3.6914 & 4.7848 & 2.8590 & 0.0170 \\ 
  s(bg\_prob\_fol) & 2.4158 & 2.7373 & 2.3562 & 0.0475 \\ 
  ti(normalized\_t,bg\_prob\_fol) & 7.8947 & 8.6462 & 7.8430 & $<$ 0.0001 \\ 
   \hline
\end{tabular}
\label{tab:model3}
\end{table}

\begin{table}[htbp]
\caption{Summary of the model fitted with \texttt{word} for the \texttt{4.0} tonal context dataset} 
\centering
\begin{tabular}{lrrrr}
   \hline
A. parametric coefficients & Estimate & Std. Error & t-value & p-value \\ 
  (Intercept) & 5.2497 & 0.0312 & 168.0817 & $<$ 0.0001 \\ 
  gendermale & -0.4725 & 0.0407 & -11.6043 & $<$ 0.0001 \\ 
   \hline
B. smooth terms & edf & Ref.df & F-value & p-value \\ 
  s(normalized\_t):genderfemale & 2.9192 & 2.9806 & 29.4234 & $<$ 0.0001 \\ 
  s(normalized\_t):gendermale & 1.0003 & 1.0005 & 5.0343 & 0.0248 \\ 
  s(speaker) & 49.0454 & 52.0000 & 30.1430 & $<$ 0.0001 \\ 
  s(normalized\_t,word) & 1251.5978 & 1890.0000 & 6.4034 & $<$ 0.0001 \\ 
  s(normalized\_t,tone\_pattern) & 93.3188 & 179.0000 & 1.4397 & $<$ 0.0001 \\ 
  s(speech\_rate):genderfemale & 2.6175 & 2.8809 & 8.3676 & 0.0001 \\ 
  s(speech\_rate):gendermale & 2.7772 & 2.9508 & 8.2388 & 0.0001 \\ 
  ti(normalized\_t,speech\_rate) & 6.8744 & 7.9913 & 6.6015 & $<$ 0.0001 \\ 
  s(norm\_utt\_pos) & 2.0931 & 2.4575 & 41.5639 & $<$ 0.0001 \\ 
  ti(normalized\_t,norm\_utt\_pos) & 7.0265 & 8.0821 & 12.4450 & $<$ 0.0001 \\ 
  s(bg\_prob\_prev) & 1.9451 & 2.3101 & 12.5304 & $<$ 0.0001 \\ 
  ti(normalized\_t,bg\_prob\_prev) & 5.9749 & 7.1845 & 4.1940 & 0.0001 \\ 
  s(bg\_prob\_fol) & 1.0006 & 1.0010 & 8.6352 & 0.0033 \\ 
  ti(normalized\_t,bg\_prob\_fol) & 5.6756 & 6.9167 & 3.1157 & 0.0029 \\ 
   \hline
\end{tabular}
\label{tab:model4}
\end{table}

\begin{table}[htbp]
\caption{Summary of the model with \texttt{sense\_type} for the \texttt{4.4} tonal context dataset} 
\centering
\begin{tabular}{lrrrr}
   \hline
A. parametric coefficients & Estimate & Std. Error & t-value & p-value \\ 
  (Intercept) & 5.2990 & 0.0255 & 207.7527 & $<$ 0.0001 \\ 
  gendermale & -0.5276 & 0.0345 & -15.2859 & $<$ 0.0001 \\ 
   \hline
B. smooth terms & edf & Ref.df & F-value & p-value \\ 
  s(normalized\_t):genderfemale & 1.0037 & 1.0046 & 22.3822 & $<$ 0.0001 \\ 
  s(normalized\_t):gendermale & 2.6128 & 2.8854 & 7.3128 & 0.0001 \\ 
  s(speaker) & 50.7591 & 53.0000 & 59.8792 & $<$ 0.0001 \\ 
  s(normalized\_t,sense\_type) & 1645.3569 & 2394.0000 & 6.1036 & $<$ 0.0001 \\ 
  s(normalized\_t,tone\_pattern) & 115.1964 & 179.0000 & 2.3487 & $<$ 0.0001 \\ 
  s(speech\_rate):genderfemale & 2.2619 & 2.6469 & 3.5374 & 0.0310 \\ 
  s(speech\_rate):gendermale & 2.2986 & 2.6760 & 8.0901 & 0.0001 \\ 
  ti(normalized\_t,speech\_rate) & 3.9985 & 4.7382 & 15.5029 & $<$ 0.0001 \\ 
  s(norm\_utt\_pos) & 2.3278 & 2.6858 & 57.3866 & $<$ 0.0001 \\ 
  ti(normalized\_t,norm\_utt\_pos) & 5.6890 & 7.1236 & 4.8475 & $<$ 0.0001 \\ 
  s(bg\_prob\_prev) & 2.8814 & 2.9813 & 24.2073 & $<$ 0.0001 \\ 
  ti(normalized\_t,bg\_prob\_prev) & 5.3450 & 6.5906 & 2.7958 & 0.0083 \\ 
  s(bg\_prob\_fol) & 1.1588 & 1.2896 & 7.5202 & 0.0028 \\ 
  ti(normalized\_t,bg\_prob\_fol) & 7.8741 & 8.6372 & 11.9355 & $<$ 0.0001 \\ 
   \hline
\end{tabular}
\label{tab:model1sense}
\end{table}

\begin{table}[htbp]
\caption{Summary of the model with \texttt{sense\_type} for the \texttt{3.4} tonal context dataset} 
\centering
\begin{tabular}{lrrrr}
   \hline
A. parametric coefficients & Estimate & Std. Error & t-value & p-value \\ 
  (Intercept) & 5.3190 & 0.0245 & 217.3049 & $<$ 0.0001 \\ 
  gendermale & -0.5279 & 0.0322 & -16.4049 & $<$ 0.0001 \\ 
   \hline
B. smooth terms & edf & Ref.df & F-value & p-value \\ 
  s(normalized\_t):genderfemale & 2.7782 & 2.9300 & 11.8593 & $<$ 0.0001 \\ 
  s(normalized\_t):gendermale & 1.0016 & 1.0022 & 0.7333 & 0.3918 \\ 
  s(speaker) & 49.3383 & 53.0000 & 23.1424 & $<$ 0.0001 \\ 
  s(normalized\_t,sense\_type) & 1311.3647 & 1980.0000 & 5.6701 & $<$ 0.0001 \\ 
  s(normalized\_t,tone\_pattern) & 90.0576 & 179.0000 & 1.2569 & $<$ 0.0001 \\ 
  s(speech\_rate):genderfemale & 2.8229 & 2.9667 & 15.2549 & $<$ 0.0001 \\ 
  s(speech\_rate):gendermale & 2.7810 & 2.9558 & 3.5426 & 0.0186 \\ 
  ti(normalized\_t,speech\_rate) & 7.5150 & 8.4879 & 7.7827 & $<$ 0.0001 \\ 
  s(norm\_utt\_pos) & 2.5743 & 2.8458 & 47.1183 & $<$ 0.0001 \\ 
  ti(normalized\_t,norm\_utt\_pos) & 7.0027 & 8.0972 & 5.2953 & $<$ 0.0001 \\ 
  s(bg\_prob\_prev) & 2.4559 & 2.7397 & 11.7554 & $<$ 0.0001 \\ 
  ti(normalized\_t,bg\_prob\_prev) & 8.5396 & 8.8648 & 32.3655 & $<$ 0.0001 \\ 
  s(bg\_prob\_fol) & 1.0011 & 1.0022 & 8.6211 & 0.0033 \\ 
  ti(normalized\_t,bg\_prob\_fol) & 2.3724 & 2.7114 & 3.1644 & 0.0175 \\ 
   \hline
\end{tabular}
\label{tab:model2sense}
\end{table}

\begin{table}[htbp]
\caption{Summary of the model with \texttt{sense\_type} for the \texttt{4.1} tonal context dataset}
\centering
\begin{tabular}{lrrrr}
   \hline
A. parametric coefficients & Estimate & Std. Error & t-value & p-value \\ 
  (Intercept) & 5.2785 & 0.0258 & 204.6860 & $<$ 0.0001 \\ 
  gendermale & -0.4733 & 0.0330 & -14.3633 & $<$ 0.0001 \\ 
   \hline
B. smooth terms & edf & Ref.df & F-value & p-value \\ 
  s(normalized\_t):genderfemale & 1.0006 & 1.0008 & 10.6364 & 0.0011 \\ 
  s(normalized\_t):gendermale & 2.2744 & 2.6385 & 4.7004 & 0.0076 \\ 
  s(speaker) & 49.3758 & 53.0000 & 25.2692 & $<$ 0.0001 \\ 
  s(normalized\_t,sense\_type) & 1390.6999 & 2052.0000 & 5.2771 & $<$ 0.0001 \\ 
  s(normalized\_t,tone\_pattern) & 89.6240 & 179.0000 & 1.2507 & $<$ 0.0001 \\ 
  s(speech\_rate):genderfemale & 1.0012 & 1.0023 & 0.8458 & 0.3574 \\ 
  s(speech\_rate):gendermale & 1.0900 & 1.1655 & 8.0534 & 0.0043 \\ 
  ti(normalized\_t,speech\_rate) & 6.2987 & 7.3658 & 11.8930 & $<$ 0.0001 \\ 
  s(norm\_utt\_pos) & 1.0007 & 1.0014 & 20.7997 & $<$ 0.0001 \\ 
  ti(normalized\_t,norm\_utt\_pos) & 4.4585 & 5.9266 & 1.7014 & 0.1187 \\ 
  s(bg\_prob\_prev) & 2.3038 & 2.6419 & 4.8417 & 0.0042 \\ 
  ti(normalized\_t,bg\_prob\_prev) & 5.4972 & 6.7456 & 3.7973 & 0.0005 \\ 
  s(bg\_prob\_fol) & 2.6706 & 2.8966 & 4.2319 & 0.0049 \\ 
  ti(normalized\_t,bg\_prob\_fol) & 7.3670 & 8.3279 & 3.9638 & 0.0001 \\ 
   \hline
\end{tabular}
\label{tab.gam}
\end{table}

\begin{table}[htbp]
\caption{Summary of the model with \texttt{sense\_type} for the \texttt{4.0} tonal context dataset} 
\centering
\begin{tabular}{lrrrr}
   \hline
A. parametric coefficients & Estimate & Std. Error & t-value & p-value \\ 
  (Intercept) & 5.2524 & 0.0327 & 160.7046 & $<$ 0.0001 \\ 
  gendermale & -0.4743 & 0.0414 & -11.4693 & $<$ 0.0001 \\ 
   \hline
B. smooth terms & edf & Ref.df & F-value & p-value \\ 
  s(normalized\_t):genderfemale & 2.9426 & 2.9856 & 31.6209 & $<$ 0.0001 \\ 
  s(normalized\_t):gendermale & 1.0016 & 1.0023 & 7.1126 & 0.0076 \\ 
  s(speaker) & 48.0016 & 53.0000 & 23.1946 & $<$ 0.0001 \\ 
  s(normalized\_t,sense\_type) & 994.9116 & 1539.0000 & 5.5151 & $<$ 0.0001 \\ 
  s(normalized\_t,tone\_pattern) & 85.9056 & 179.0000 & 1.1669 & $<$ 0.0001 \\ 
  s(speech\_rate):genderfemale & 2.0016 & 2.3892 & 2.3245 & 0.1149 \\ 
  s(speech\_rate):gendermale & 2.1820 & 2.5488 & 4.7352 & 0.0061 \\ 
  ti(normalized\_t,speech\_rate) & 7.1264 & 8.2127 & 7.6463 & $<$ 0.0001 \\ 
  s(norm\_utt\_pos) & 1.8211 & 2.1735 & 46.3419 & $<$ 0.0001 \\ 
  ti(normalized\_t,norm\_utt\_pos) & 6.9596 & 8.0649 & 10.2243 & $<$ 0.0001 \\ 
  s(bg\_prob\_prev) & 2.1932 & 2.5430 & 13.3047 & $<$ 0.0001 \\ 
  ti(normalized\_t,bg\_prob\_prev) & 6.8339 & 7.8695 & 6.2328 & $<$ 0.0001 \\ 
  s(bg\_prob\_fol) & 1.0194 & 1.0344 & 7.0325 & 0.0073 \\ 
  ti(normalized\_t,bg\_prob\_fol) & 1.9840 & 2.6243 & 0.4195 & 0.7027 \\ 
   \hline
\end{tabular}
\label{tab.gam}
\end{table}

\clearpage
\section*{Appendix 2: the effects of segments and frequency} 

\noindent
Our findings demonstrate that the word itself is a strong predictor of pitch contours in disyllabic words. However, one might question whether this robust effect is at least partially influenced by segmental properties, given existing evidence on the impact of vowel height and onset consonants on Mandarin tones \citep{ho1976acoustic, ladd1984vowel, ohala1976explaining, whalen1995universality}. Additionally, lexical frequency has long been recognized as a factor influencing f0 contours, with lower-frequency words being produced with higher pitch \citep{zhao2007effect}. To address these concerns, this additional analysis clarifies the effects of word's segmental composition and lexical frequency on pitch contours.

Following \citet{chuang2024word}, for our disyllabic words, we coded four predictors for segments. 
\texttt{vowel1\_height} and \texttt{vowel2\_height} are the vowel height of the first syllable and the second syllable, respectively. Each has five levels: `high', mid', and low', low-high' and mid-high'.
\texttt{onset1\_type} and \texttt{onset2\_type} are the type of the onset consonant of the first syllable and the second syllable, respectively. Each has seven levels: `aspirated-affricate', `aspirated-stop', `unaspirated-affricate', `unaspirated-stop', `voiceless-fricative', `voiced', and `null'. 
\texttt{frequency} represents the log-transformed count of occurrences of a word type in entire spoken corpus of Taiwan Mandarin.

We built up a baseline model that includes gender, tonal context, tone pattern, speaking rate, speaker, word position, bigram probability, but excludes word. To simplify the analysis, this model was based on an omnibus dataset that integrates all four tonal contexts.

To examine the effects of segments, four factor smooth terms for \texttt{vowel1\_height}, \texttt{vowel2\_height}, \texttt{vowel1\_type}, and \texttt{vowel2\_type} were added to the baseline model together. 
\begin{tabbing}
mmmmm\=mm\=\kill
baseline +
       \> \> \texttt{s(normalized\_t, vowel1\_height, bs=`fs', m=1)}+ \\
       \> \> \texttt{s(normalized\_t, vowel2\_height, bs=`fs', m=1)}+ \\
       \> \> \texttt{s(normalized\_t, onset1\_type, bs=`fs', m=1)}+ \\
       \> \> \texttt{s(normalized\_t, onset2\_type, bs=`fs', m=1)}  \\
\end{tabbing}

To examine the effect of frequency, we added the smooth term for \texttt{frequency}, in combination with its interaction with \texttt{normalized\_t}, to the baseline model. 

\begin{tabbing}
mmmmm\=mm\=\kill
baseline +
       \> \> \texttt{s(frequency, k=4)+} \\
       \> \> \texttt{ti(normalized\_t, frequency, k=c(4,4))}
\end{tabbing}

Table~\ref{tab:segments} presents the improvement of model fit compared with the baseline model, as evaluated by AIC change. The inclusion of the four segmental predictors combined improves the model fit by 5,267.08 units, while the inclusion of \texttt{word} leads to a more substantial improvement of 21,532.92 units. Moreover, in \citet{chuang2024word}, with only around 50 word types, the segment-related controls were highly confounded with one another, as indicated by the high concurvity scores of around 0.75. However, in our dataset that contains a greatly larger number of word types, these effects can be better disentangled. For the \texttt{baseline + four segmental predictors} model, the concurvity scores are much lower than those reported in \citet{chuang2024word}: 0.42 for \texttt{s(normalized\_t,vowel1\linebreak\_height)}, 0.40 for \texttt{s(normalized\_t,vowel2\_height)}, 0.35 for \texttt{s(normalized\_t,onset1\_type)}, and 0.35 for \texttt{s(normalized\_t,onset2\_type)}. In the baseline + \texttt{word} model, the concurvity score for \texttt{word} is also low (0.12). 
Additionally, although \texttt{baseline + four segmental predictors + word} has the lowest AIC, concurvity of four segmental predictors is all 1, and that of \texttt{word} is 0.21, suggesting that segmental predictors are highly colinear with \texttt{word} when they are both present.

\begin{table}[htbp]
\caption{Improvement of model fit gauged by AIC change}
\centering
\adjustbox{max width=\textwidth}{
\begin{tabular}{cccc}
  \hline
  \textbf{Model} & \textbf{AIC} & \textbf{AIC difference} \\ 
  \hline
  \texttt{baseline} & 279.70 & -637646.74 & - \\ 
  \texttt{baseline + frequency} & 291.03 & -637852.51 & -205.76 \\ 
  \texttt{baseline + four segmental predictors} & 447.63 & -642913.82 & -5267.08 \\ 
  \texttt{baseline + word} & 2475.79 & -659179.66 & -21532.92 \\ 
  \texttt{baseline + frequency + word} & 2475.54 & -659189.68 & -21542.93 \\ 
  \texttt{baseline + four segmental predictors} & 2444.84 & -659198.24 & -21551.50 \\ 
   \hline
\end{tabular}}
\label{tab:segments}
\end{table}

Similarly, the inclusion of \texttt{word} (21532.92 AIC units) resulted in a more substantial AIC decrease than \texttt{frequency} (205.76 AIC units). Besides, we added \texttt{frequency} on top of the \texttt{baseline + word} model, which resulted in a further AIC decrease of 10.01 units. However, in this model, concurvity was very high for \texttt{frequency} (0.99) and low for \texttt{word} (0.13).

Overall, \texttt{word} by itself contributes more to the model fit than the four segmental predictors combined, as well as lexical frequency. In line with \citet{chuang2024word}, the effect of word cannot be simply reduced to the effect of segments. Besides, word is a better predictor of pitch contours than lexical frequency.

\clearpage
\normalsize
\bibliography{Ref}

\end{CJK*}
\end{document}